\pdfoutput=1
\documentclass[11pt]{article}

\usepackage[utf8]{inputenc}
\usepackage[T1]{fontenc}
\usepackage{lmodern}
\usepackage[margin=1in]{geometry}
\usepackage{microtype}

\usepackage{amsmath,amssymb}
\usepackage{newunicodechar}
\newunicodechar{→}{\ensuremath{\rightarrow}}
\newunicodechar{↔}{\ensuremath{\leftrightarrow}}
\newunicodechar{−}{\ensuremath{-}}
\newunicodechar{≈}{\ensuremath{\approx}}
\newunicodechar{≥}{\ensuremath{\geq}}
\newunicodechar{×}{\ensuremath{\times}}

\usepackage{booktabs,longtable,array,calc,tabularx}
\providecommand{\tightlist}{%
  \setlength{\itemsep}{0pt}\setlength{\parskip}{0pt}}
\providecommand{\real}[1]{#1}

\usepackage{graphicx,xcolor}
\usepackage{tikz}
\usetikzlibrary{arrows.meta,positioning,shapes.geometric,calc,decorations.pathreplacing}
\usepackage{chessboard}
\usepackage{flafter}

\usepackage[hidelinks]{hyperref}
\hypersetup{
  pdftitle={Zugwise: Domain-Aware Game Reconstruction for Digitizing Handwritten Chess Scoresheets},
  pdfauthor={Gerhard Trippen},
  pdfsubject={Handwritten chess scoresheet digitization},
  pdfkeywords={handwriting recognition, chess, game reconstruction, human-in-the-loop, quiescence search}}

\usepackage[nameinlink]{cleveref}
\crefformat{section}{\S#2#1#3}
\crefformat{subsection}{\S#2#1#3}
\crefformat{subsubsection}{\S#2#1#3}
\Crefformat{section}{\S#2#1#3}
\Crefformat{subsection}{\S#2#1#3}
\Crefformat{subsubsection}{\S#2#1#3}
\crefname{figure}{Figure}{Figures}
\Crefname{figure}{Figure}{Figures}
\crefname{table}{Table}{Tables}
\Crefname{table}{Table}{Tables}

\title{\textbf{Zugwise: Domain-Aware Game Reconstruction for\\
Digitizing Handwritten Chess Scoresheets}}
\author{Gerhard Trippen\\[2pt]
  \normalsize Rotman School of Management, University of Toronto}
\date{Preprint --- October 2026}

\begin{document}
\maketitle

\begin{abstract}
\noindent
Digitizing handwritten chess scoresheets has been studied as a
handwriting-recognition problem: read the move in each box and check that it
is legal. This framing misses the errors that matter most. A misread that
yields an \emph{illegal} move is loud. One that yields a
\emph{legal-but-wrong} move is silent, because a legality-only pipeline
accepts it and reads every later move against a wrong position. We therefore
treat the task as \emph{game reconstruction} and present \emph{Zugwise},
which repurposes computer-chess machinery (quiescence search, static exchange
evaluation, pin detection) to ask not ``what should be played?'' but ``could a
human plausibly have written this?'' An absurdity detector makes silent errors
loud at or near where they occur, and a backward search traces late symptoms
to their cause. Candidate ranking excludes chess-quality signals. A
Needleman--Wunsch merge adds the second player's sheet as evidence, and review
stops send every disagreement between the sheets' legal readings to a human.
In oracle replay of two tournament sections, a release adding them cuts
silent errors from 16 to 1 and from 50 to 13. Two learned
signals, CTC re-scoring against the recognizer's raw output and a Maia-style
human-move prior, raise the correct fix's top-rank rate from 58.8\% to 78.0\%
of the replacement decisions where it is a candidate, and hold without
refitting on a later section. On the first section,
the human-verified output matches an independent expert transcription move for
move on 34 of 45 games, and adjudication against the sheets found errors in
both records, on different plies. We do not establish that
either workflow is faster or more accurate. The system runs entirely in the
browser and is free, private, and offline-capable; alongside it we offer a
three-way adjudication protocol for a domain where no single record is ground
truth.
\end{abstract}

\section{Introduction}\label{sec:intro}

Over-the-board tournament chess still runs on paper. The FIDE Laws of
Chess~\cite{fide} require players to record the moves of standard-play games,
so most over-the-board classical games begin their existence as a pair of
handwritten scoresheets, and very few of them end up anywhere else. A modest
120-player weekend tournament of five rounds produces some three hundred
games and at least six hundred handwritten sheets, since a long game can fill
more than one sheet per player. Typing them up is laborious enough that,
outside elite events, it is mostly not done at all. The games are not lost;
they sit in boxes.

Research on digitizing these scoresheets has, naturally, treated it as a
handwriting recognition problem. A pipeline segments the sheet into move
boxes, recognizes the text in each box, and checks that the result is a legal
move (\cref{sec:related}). That program has been carried impressively far.
The best published systems cross-check the two players' scoresheets and
recognize about 95\% of moves without human input, and about 98--99\% when a
person resolves the flagged cases~\cite{eicher2021,majid2022}. This paper
begins where that objective saturates, with the errors recognition cannot
see.

A recognition error that produces an \emph{illegal} move is loud. Every
system since the earliest legality checkers catches it, and it costs only a
correction. A recognition error that produces a \emph{legal} move is silent.
A legality-only pipeline accepts it, advances the board to a wrong position,
and then interprets every later move against that position. We call this
\textbf{cascading corruption}. An earlier system
ran into it independently: it attempted iterative game-state tracking and
abandoned it when a single wrong move corrupted everything
downstream~\cite{horvath2020}. Majid and Eicher note that a valid but wrongly
written move may surface only later, as a contradiction that makes a true
move look illegal~\cite{majid2022}. High per-move accuracy therefore does not
certify a correct game. Pipelines that correct illegal readings but accept
legal ones let the legal-but-wrong readings escape review. Confidence checks
catch the unconfident ones, but a confident legal-but-wrong reading passes
every character-level and validity check (\cref{sec:eval:measure}). Even at
99\% per-move accuracy, an 80-ply game expects nearly one erroneous move.
Whatever share of games that affects, the silent errors among them leave an
archive with corrupted games and nothing to flag which ones to distrust.

This paper therefore argues for a reframing. Scoresheet digitization is not a
recognition problem with a legality check appended but a \textbf{game
reconstruction} problem. The object to be recovered is the game, and the
scoresheet is noisy evidence about it. For every reading, the question is no
longer ``what character string is in this box?'' but ``what sequence of moves
was actually played, given this evidence?'' To answer it, we borrow the
machinery of computer chess (quiescence search, static exchange evaluation,
pin detection) and turn it to a question engines never ask. An engine asks
\emph{what should be played}; we ask \emph{could a human plausibly have
written this}. Consider a legal position in which a queen stands freely
capturable for three consecutive moves while both players ignore it. It is
rarely a record of human play, and more often a record of an OCR error.
Detecting such \emph{positional absurdities} turns silent errors into loud
ones, a conversion a legality check cannot perform.

Plausibility, however, has a boundary, and this paper is about where it
lies. In one game of our corpus, plausibility reasoning resolves an
ambiguity that no character-level analysis could. Two readings are both
legal, and only one of them hands the opponent a free capture with check. In
another game from the same tournament, the same kind of reasoning points the
wrong way. A player deliberately captured a queen and allowed the mate that
followed, as a way of resigning. A position-only plausibility heuristic would
reject this \emph{resignation capture}. What separates the two cases lies not
in the position but in the rest of the record and in the handwriting
(\cref{sec:recon:matchedpair}, \cref{sec:disc:helps}). For this reason the
system is built around a human operator rather than in spite of one. Its
scoring follows a design rule. Signals about what \emph{should} be played are
excluded from candidate ranking, and among hand-built chess signals only
plausibility evidence is admitted, fallible but checked by tactical search
(\cref{sec:recon:scoring}). Added learned and engine-derived ranking terms must
also meet an admission standard: at settings fixed in advance, they must
improve ranking on games not used to choose them, and their effect on
correctly recorded blunders is reported. The rule does not exclude knowledge of how people
actually play. A human-move prior in the manner of
Maia~\cite{mcilroyyoung2020maia}, trained on some 8.4 million positions from
120{,}000 online games, estimates how often players in its training band make
each move. It is an estimate learned from the population's games, not a
hand-written opinion about the position. It also answers the reconstruction
question directly: of the moves the ink could be, which would a club player
most likely have made? Combined with re-scoring every candidate against the
recognizer's raw output, it is the largest ranking gain we measured
(\cref{sec:recon:learned}, \cref{sec:eval:learned}). The unchanged prior meets
that standard on a section digitized after its weight was fixed. CTC also
gains there without refitting, although its two-sheet form was introduced
while that section was being digitized. A bounded engine evaluation, tested
the same way, was not confirmed (\cref{sec:eval:engine}). The system's job is to
make the operator's decisions few and sharp.

We present \textbf{Zugwise},\footnote{Freely available, in-browser, at
\url{https://gerhardtrippen.github.io/Zugwise/}~\cite{zugwiserepo}.} an implementation of
this approach that runs entirely in the browser. A phone photograph or scan
of a scoresheet first passes through grid detection, which anchors on the
printed move-number columns, assigns each row its move number with a sliding
window over the holes in the printed digits, and extracts every cell with its
own small perspective correction (\cref{sec:arch:grid}). It then
goes into a BiLSTM recognizer built directly on the architecture and dataset
of Eicher et al.~\cite{eicher2021,majid2022}, and from there into the
reconstruction and verification layers that are this paper's subject.
Everything runs client-side (Pyodide and ONNX Runtime Web). The tool is free,
private, and offline-capable, and it costs nothing to operate. We chose these
properties deliberately: the scoresheets in question belong to clubs and
children's tournaments.

We evaluate where the silent-error framing demands, at the game level and
against the realistic alternative. A complete 45-game round-robin section of
a rated tournament was processed twice. Zugwise processed it with human
verification, and an experienced transcriber independently typed the same
scoresheets by hand. Every disagreement was then adjudicated against the
original sheets, because neither record is ground truth for the other. The
comparison itself caught a queen left hanging in the expert's record and
king-walk squares misread by Zugwise. Three quarters of the games (34 of 45)
come out move-for-move identical. We do not establish that either workflow is
faster or more accurate. Among the adjudicated disagreements, the two
processes err on \emph{different} plies.
Both err on ambiguous squares and routes in chess-permissive positions; only
hand transcription errs on whether the hardest, time-trouble-degraded moves
get recorded at all. At no adjudicated disagreement were both records wrong,
so each catches errors the other contains (\cref{sec:eval:taxonomy}). Because
agreeing moves were not independently audited, errors the two records share
cannot be excluded. Where no single record is ground truth, we offer the
three-way adjudication protocol as a contribution.

\medskip
\noindent\textbf{Contributions.}

\begin{enumerate}
\def\labelenumi{\arabic{enumi}.}
\tightlist
\item
  A reframing of scoresheet digitization as \textbf{game reconstruction}
  rather than document analysis, locating the binding failure mode
  (silent, legal-but-wrong readings and the cascading corruption they
  cause) above the recognition layer.
\item
  A \textbf{positional absurdity detector}, to our knowledge the first
  published for scoresheet reconstruction, that repurposes
  quiescence search to flag legal-but-implausible move sequences as
  OCR-error signals.
\item
  A unified candidate-scoring function governed by a \textbf{design rule
  excluding chess-quality signals}, admitting among hand-built chess
  signals only plausibility evidence checked by tactical search, together with a test that separates a
  hand-written judgment of the position from a permitted \textbf{learned
  human-move prior}. Combined with re-scoring of candidates against the
  raw recognizer output, the prior raises the top-rank rate of the correct
  fix from 58.8\% to 78.0\% of the replacement decisions at which the
  correct move is a candidate, and keeps its gain without refitting on a
  section digitized later.
\item
  A \textbf{layered error-correction architecture} (instant checks,
  character-confusion modeling, tactical-inconsistency detection, and
  frontier-based backtracking) with a validation-frontier mechanism for
  interactive, human-in-the-loop correction. Review stops send every
  disagreement between the sheets' legal readings to the operator and
  re-check what has been confirmed; in oracle replay of a 45-game section
  a release that adds them and other reconstruction changes leaves one silent
  error where an earlier release leaves 16 (\cref{sec:eval:ablation}). The layers are complementary.
  Absurdity detection stops the replay at the error, and when an error
  surfaces only plies later, the backward search traces the symptom back to
  its cause. With the detector disabled, the search still offered the
  correct earlier fix for every such error in our corpus, from a stop a
  median of two plies away (\cref{sec:eval:ablation}).
\item
  A dual-scoresheet merging method adapting \textbf{Needleman--Wunsch
  sequence alignment} with OCR candidate-pool overlap scores.
\item
  A robust \textbf{grid-detection method} that anchors on the printed
  move-number columns and assigns move numbers to rows with a sliding
  window over the holes in the printed digits, tolerating missing or
  overwritten numbers. Commercial systems
  solve this stage with proprietary, unpublished methods.
\item
  A \textbf{three-way adjudication protocol}, and the empirical lessons
  from applying it, for evaluating transcription systems in a domain where
  no single record, human or machine, is ground truth.
\item
  A fully client-side, free, private, offline-capable deployment,
  demonstrating that the entire pipeline fits in a browser.
\end{enumerate}

\medskip
\noindent\textbf{Terminology.} A \emph{ply} is one player's move; where
the paper says ``a move'' without a number, it means one ply. A
\emph{move number} covers White's ply and Black's reply: move 25 consists
of 25.W and 25.B. When plies are counted, they are counted from zero, so
25.W is ply 48 and 25.B is ply 49. A \emph{move box} is the scoresheet
cell holding one ply, so per-move recognition accuracy is per box.
\emph{Recognition} means reading the text in a box; \emph{reconstruction}
means recovering the game. Two records \emph{agree} when they contain the
same move at a ply; agreement is not correctness, and \emph{correct}
means supported by the original sheets on adjudication.

The remainder of the paper proceeds as follows. \cref{sec:related} surveys prior work
and locates the gap. \cref{sec:arch} describes the system architecture from
photograph or scan to verified PGN. \cref{sec:recon} develops the reconstruction
layer and is the heart of the paper: absurdity detection, the plausibility
boundary, the scoring design rule, the two learned signals, fix-finding, and
dual-sheet alignment. \cref{sec:eval} evaluates at the game level against the
independent expert transcription, including an ablation, the learned
signals' effect on ranking, and a legality-only baseline. \cref{sec:disc}
discusses what the matched pair implies about the limits of plausibility
reasoning, and the system's limitations; \cref{sec:concl} concludes.

\section{Related Work}\label{sec:related}

\subsection{Chess Scoresheet Recognition}\label{sec:related:scoresheet}

Academic work on this problem is surprisingly thin given its practical
demand. We survey it chronologically (\cref{tab:relatedwork}).\footnote{Several ZHAW Winterthur student works
\cite{horvath2020,caglayan2021,hostettler2023,abduli2020,ambrosini2022,hostettler2022sem} share institutional supervision and form a
continuous line of work.}

\begin{longtable}[]{@{}
  >{\raggedright\arraybackslash}p{(\columnwidth - 8\tabcolsep) * \real{0.14}}
  >{\raggedright\arraybackslash}p{(\columnwidth - 8\tabcolsep) * \real{0.22}}
  >{\raggedright\arraybackslash}p{(\columnwidth - 8\tabcolsep) * \real{0.15}}
  >{\raggedright\arraybackslash}p{(\columnwidth - 8\tabcolsep) * \real{0.17}}
  >{\raggedright\arraybackslash}p{(\columnwidth - 8\tabcolsep) * \real{0.32}}@{}}
\caption{Academic work on handwritten chess scoresheet recognition, in
chronological order.}\label{tab:relatedwork}\\
\toprule
Ref & Authors & Year & Venue & Contribution \\
\midrule
\endfirsthead
\toprule
Ref & Authors & Year & Venue & Contribution \\
\midrule
\endhead
\cite{baird1990} & Baird \& Thompson & 1990 & IEEE TPAMI & Reading the printed
game text of a chess encyclopedia (not handwriting), using the rules of
chess to check and correct the recognized moves --- an early precedent
for game-level semantic correction of OCR \\
\cite{kochanska2017} & Kochańska & 2017 & PhD thesis, Warsaw U. of Tech. & Thesis
on handwritten chess scoresheets, cited by~\cite{majid2022}. Listed for
completeness; we could not obtain a copy to verify its content \\
\cite{horvath2020} & Horváth \& Dreher & 2020 & BSc thesis, ZHAW Winterthur &
``ChessReader.'' Compared three commercial handwriting-OCR services
(ABBYY, LEADTOOLS, Google Cloud Vision) on features and price and built
on ABBYY's cloud OCR; prototyped and set aside a custom CNN approach
(MSER/EMNIST) because handwritten characters overlap; used Tesseract only
to locate printed move numbers. Published a confusion matrix of ABBYY's
recognition errors, which we draw on for similarity calibration. \textbf{Attempted iterative game-state tracking
and abandoned it when one wrong move corrupted all subsequent positions
--- an early, independent discovery of what we term cascading
corruption.} \\
\cite{eicher2021} & Eicher, Farmer, Li \& Majid & 2021 & ICDAR Workshop (CBDAR) &
First peer-reviewed paper on handwritten scoresheets. BiLSTM+CTC; HCS
dataset (158 games). Raw 82.5\% MRA; 90.1\% autonomous and 97.2\%
semi-autonomous, both post-processing schemes cross-checking the two
players' sheets. Its conclusion sketches, as future work, board-tracking
post-processing, a search for the fewest corrections yielding a valid
game, and a most-played-moves prior \\
\cite{caglayan2021} & Caglayan \& Nielsen & 2021 & BSc thesis, ZHAW Winterthur &
Character-level ensembling approach; continuation of the ZHAW line \\
\cite{majid2022} & Majid \& Eicher & 2022 & J. Imaging (MDPI) & Extended journal
version: more training data, a rare-move set and Stockfish legality
checking. Default pipeline 87.8\% raw, 94.9\% autonomous, 98.4\%
semi-autonomous (95.6\% and 99.1\% with manual four-corner marking).
Same future-work sketch as~\cite{eicher2021} \\
\cite{hayashi2022} & Hayashi \& Hirata & 2022/2024 & IEEE ICPR / arXiv & Attention
encoder-decoder case study on the HCS dataset. Focus on understanding
network attention, not a deployable system \\
\cite{hostettler2023} & Hostettler \& Boner & 2023 & BSc thesis, ZHAW Winterthur &
``ChessReader Corpus'' --- a labeled multilingual scoresheet corpus
(62 games, roughly 4{,}000 moves; English/German/French notation) built
on the ZHAW pipeline \\
\cite{zhu2024} & Zhu & 2024 & IEEE MIT URTC & Student paper. BiLSTM; 94.36\%
CRA general, 95.22\% personalized. Basic move-validity checker \\
\cite{kostka2025} & Kostka \& Fischer & 2025 & GI (AKWI) & TrOCR web app
(microservices). \textasciitilde45\% CER --- suggests that
general-purpose encoder-decoder architectures struggle without
domain-specific adaptation \\
\bottomrule
\end{longtable}

\textbf{Critical observation.} Using the rules of chess to correct
recognition is not new: Baird and Thompson did so for printed game
text~\cite{baird1990}. Among the handwritten-scoresheet works we could examine,
however, all stop at character/move recognition plus cross-sheet and
legality checks. None performs game-level reconstruction, although
Eicher et al.\ name it as the next step. Both their papers sketch
post-processing over whole-game paths rather than move by move: a search
for the fewest corrections that yield a valid game, and a prior over the
most-played moves~\cite{eicher2021,majid2022}. This paper takes up that
direction, with one change of objective that \cref{sec:eval:priorwork}
motivates. A valid game is not enough, so the search is scored by
plausibility evidence rather than by the number of corrections alone.
The semi-autonomous mode of~\cite{eicher2021,majid2022} prompts the user on
invalid readings, on low-confidence readings and on conflicts between
the two sheets, and so catches many errors. What it accepts silently is
a reading that is valid and confident on at least one sheet, unless the
other sheet confidently disagrees. A legal-but-wrong reading passes in
this case. The difference is architectural. Prior pipelines apply chess
knowledge at the \emph{recognition} stage, using legality as a post-hoc
filter on character reads. We apply it at the \emph{reconstruction}
stage, asking whether a coherent game is consistent with the reads.

\textbf{Commercial tools.} Several applications exist for scoresheet
digitization, and they span a spectrum of game-awareness. Some confirm
every move with the user. Others flag low-confidence readings (e.g.,
color-coding how well a move fits the OCR result, or halting when
confidence drops too low). At least one attempts to construct a
coherent game, and some now also flag individual moves as implausible.
We characterize this spectrum without naming specific products. Our
observations are of services as they behaved when we tested them, which
for a commercial product is a moving target.

Search over alternative readings is therefore not what distinguishes
this work. At least one system appears to search. Given both players'
sheets for a game, it returns a complete legal game that is not the
concatenation of either sheet's top reads. A pipeline cross-checking the
two sheets could also produce such a combination, but it suggests a
search over readings. Two things differ. The first is \emph{what the
search optimizes}. A search scored by legality and recognition
confidence alone has no way to prefer the plausible legal move to the
implausible one. It settles cheerfully on positions a player would not
have produced, which is the same ceiling the academic line reaches,
approached from the other direction. The second, and the one the
operator feels, is \emph{what happens at the point of doubt}. These
systems commit to a reading and present the result as the game, leaving
the user to find the error by replaying it. The alternative developed
here stops at the doubt and searches backward for the earlier misreading
that caused it. It then puts ranked candidates in front of a human who
decides (\cref{sec:recon:fixfinding}).

\textbf{Why not a large language model?} General-purpose vision-language
models are an obvious alternative front end, and some commercial offerings
now advertise their use. They may well read an individual move box more
accurately than a purpose-trained recognizer. In the early tests
described below, the strongest model read more moves correctly than our
recognizer's raw top read does on club sheets (\cref{sec:arch:ocr}),
though the sheets differ. We did not adopt one, for reasons that follow
from the framing of this paper rather than from any objection to the
models themselves. First, the reconstruction layer depends on evidence
it can audit: a ranked list of candidates per box and, for CTC
re-scoring, the recognizer's raw per-frame output (\cref{sec:arch:ocr},
\cref{sec:recon:learned}). A generative model returns a single fluent
string, and where the ink is unclear it may substitute a plausible move
for the one actually written. Such output is reconstruction without an
audit trail, and it produces the silent, legal-but-wrong readings this
paper sets out to surface. Second, current language models remain
unreliable at maintaining an exact board state across a complete game,
a task an exact move generator solves outright. We therefore delegate
legality and board state to \texttt{python-chess}~\cite{pythonchess} and
use learned models only where they report measurable quantities. Third
are the deployment constraints of \cref{sec:arch:deploy}. Sending
scoresheet images, many from junior events, to a hosted service
conflicts with a design in which no image leaves the device. A per-game
fee conflicts with archiving whole tournaments at no marginal cost. And
a hosted model that is updated or retired beneath the tool makes results
difficult to reproduce. An early prototype of this work did use hosted
vision models for recognition (December 2025 to January 2026). The
strongest model tested, Claude Opus 4.5, read 91--96\% of moves
correctly on two test sheets at roughly US\$0.05--0.09 per scoresheet
(some US\$30--55 for the 600 sheets of the weekend tournament of
\cref{sec:intro}). A cheaper model of the same family cost about half as
much and read only 68--82\%. The cost is per token and dominated by the output, so it grows
with the length of the game. The image's share is set by its pixel
dimensions, and the service downscales large images before processing,
so scanning beyond that size adds no cost. The offline recognizer replaced these calls. None
of this rules out language models as a further source of evidence,
provided their readings enter the candidate pool and are scored there.

\subsection{Handwritten Text Recognition}\label{sec:related:htr}

\begin{itemize}
\tightlist
\item
  \textbf{BiLSTM + CTC:} Graves et al.~(2006)~\cite{graves2006} introduced CTC
  loss; Shi, Bai \& Yao (2017)~\cite{shi2017} the CRNN architecture. The OCR
  component we build upon (Eicher et al.'s BiLSTM) uses this paradigm
  with a constrained chess vocabulary (ours has 27 characters and the CTC
  blank, \cref{sec:recon:learned}).
\item
  \textbf{TrOCR:} Li et al.~(2023)~\cite{li2022trocr}, Transformer-based OCR.
  Kostka and Fischer~\cite{kostka2025} attempted this for scoresheets with poor results
  (\textasciitilde45\% CER), suggesting the constrained-vocabulary CTC
  approach remains better suited to this domain.
\item
  \textbf{IAM Handwriting Dataset:} Marti \& Bunke (2002)~\cite{marti2002},
  the standard pretraining source used by~\cite{eicher2021,majid2022,zhu2024}.
\end{itemize}

\subsection{Techniques Borrowed from Computer Chess}\label{sec:related:chess}

A key novelty is repurposing computer-chess techniques for a
fundamentally different end:

\begin{itemize}
\tightlist
\item
  \textbf{Quiescence search:} Shannon (1950)~\cite{shannon1950} introduced the concept;
  every modern engine uses it to avoid evaluating positions
  mid-tactical-sequence. We use it to determine whether a piece is
  \emph{genuinely} hanging (absurdity) or only \emph{appears} hanging
  due to tactical complications (trap, pin, sacrifice).
\item
  \textbf{Static Exchange Evaluation (SEE):} standard for estimating
  capture sequences on one square. We use it in the bad-trade check of
  absurdity detection: does a move place a piece where the exchange on
  that square loses it material (\cref{sec:recon:absurdity})?
\item
  \textbf{Pin detection:} identifying absolute pins. Used to prevent
  false absurdity signals.
\item
  \textbf{Human move prediction:} Maia~\cite{mcilroyyoung2020maia} showed
  that a policy network trained on human games at a given rating level
  predicts the moves humans actually make, rather than the best ones. We
  follow that idea with a network of our own (eight residual blocks of
  128 filters, against Maia's six of 64), trained on the lichess open
  database~\cite{lichessdb}. We use it as a prior over what a scoresheet
  records (\cref{sec:recon:learned}). Two successors point where such a
  prior can go. Maia-2~\cite{tang2024maia2} conditions one model on
  playing strength instead of training a separate model per rating band,
  and behavioral stylometry~\cite{mcilroyyoung2021stylometry} identifies
  an individual player from their moves alone. A scoresheet is one named
  writer's record of both players' moves, at known ratings, so both are
  unusually well matched to this setting (\cref{sec:disc:future}).
\end{itemize}

\textbf{The distinction we draw:} engines use these techniques to answer
``what should you play?''; we use them to answer ``could a human have
plausibly played this?''

\subsection{Sequence Alignment}\label{sec:related:align}

\begin{itemize}
\tightlist
\item
  \textbf{Needleman--Wunsch} (Needleman \& Wunsch, 1970)~\cite{needleman1970}: global
  sequence alignment, originally for protein comparison. We adapt it for
  dual-scoresheet alignment, replacing the amino-acid substitution
  matrix with OCR candidate-pool overlap scores. The alignment naturally
  detects shift errors (one player's move numbers offset from the
  other's) and isolates regions of genuine disagreement requiring human
  adjudication.
\item
  \textbf{Durbin, Eddy, Krogh \& Mitchison (1998)}~\cite{durbin1998}: \emph{Biological
  Sequence Analysis}, the textbook reference for the DP algorithm. To
  our knowledge, applying biological sequence-alignment algorithms to
  chess scoresheet merging has not been described before.
\end{itemize}

\subsection{The HCS Dataset}\label{sec:related:hcs}

The Handwritten Chess Scoresheet (HCS) dataset~\cite{hcsdataset}, created by Eicher et
al.~\cite{eicher2021,majid2022}, is to our knowledge the largest publicly available
labeled dataset for this domain. It holds 158 games across 215
scoresheet pages as described in~\cite{eicher2021} (the journal
version~\cite{majid2022} reports 211), with about 13,810 handwritten
moves among 25,327 extracted move-box images~\cite{majid2022}. A
preprocessed version is available on Hugging Face~\cite{hcshf}. The ZHAW
ChessReader Corpus~\cite{hostettler2023} (62 games, CC BY 4.0) is
smaller and multilingual. Commercial systems presumably train on their
own data, which is not published. We train our BiLSTM on HCS. Its
ground-truth PGNs also provide a potential end-to-end benchmark, though
the dataset represents a limited range of formats and handwriting
styles. For broader coverage, our evaluation (\cref{sec:eval})
supplements HCS-derived results with real games from a club tournament
(player names withheld).

\section{System Architecture}\label{sec:arch}

This section describes the end-to-end system that turns a photograph or
scan of a scoresheet into a PGN game. The contribution of this paper is the
reconstruction layer (\cref{sec:recon}). Here we lay out the pipeline around
it and state which parts are our own work and which we build upon. In
particular, \textbf{the OCR layer is built on the BiLSTM
recognizer of Eicher et al.~\cite{eicher2021,majid2022}}. Our novelty begins at the move
candidates it produces.

\subsection{Pipeline overview}\label{sec:arch:pipeline}

\begin{figure}[t]
\centering
\resizebox{\textwidth}{!}{%
\begin{tikzpicture}[
  node distance=6mm,
  stage/.style={draw, rounded corners, align=center, minimum height=12mm,
                text width=20mm, font=\small, fill=blue!4},
  io/.style={draw, align=center, minimum height=12mm, text width=16mm,
             font=\small, fill=black!5},
  arr/.style={-{Stealth[length=2.5mm]}, thick}]
  \node[io]    (photo) {Photo / scan};
  \node[stage, right=of photo] (grid)  {Grid\\ detection};
  \node[stage, right=of grid]  (cell)  {Per-cell\\ extraction};
  \node[stage, right=of cell]  (ocr)   {BiLSTM / ONNX OCR};
  \node[stage, right=of ocr]   (recon) {Recon-\\struction};
  \node[stage, right=of recon] (verif) {Human\\ verification};
  \node[io,    right=of verif] (pgn)   {PGN};
  \foreach \a/\b in {photo/grid, grid/cell, cell/ocr, ocr/recon, recon/verif, verif/pgn}
    \draw[arr] (\a) -- (\b);
  \node[font=\scriptsize\itshape, below=1mm of ocr, text width=24mm, align=center]
        {ranked candidate\\ lists, not a\\ single string};
  \draw[decorate, decoration={brace, amplitude=4pt}, thick]
        ([yshift=5mm]grid.north west) -- ([yshift=5mm]recon.north east)
        node[midway, above=4pt, font=\scriptsize] {stages 1--3 run \emph{ahead} of the operator};
\end{tikzpicture}}
\caption{End-to-end pipeline. A photograph or scan of a scoresheet passes through grid
detection, per-cell extraction, and BiLSTM/ONNX recognition into the
reconstruction and human-verification layers, emitting standard PGN. The OCR
layer carries a ranked candidate list per cell rather than a single read, which
is what makes downstream reconstruction possible; stages~1--3 are computed
eagerly so the operator's time is spent only on moves the system could not
settle on its own.}
\label{fig:pipeline}
\end{figure}
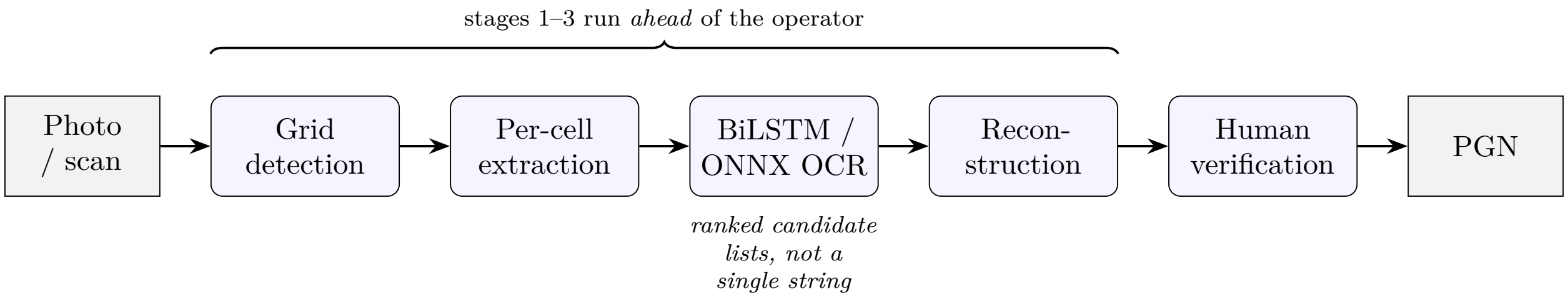

A single game flows through five stages (\cref{fig:pipeline}), all
of which run on the user's own device:

\begin{enumerate}
\def\labelenumi{\arabic{enumi}.}
\tightlist
\item
  \textbf{Grid detection} (\cref{sec:arch:grid}). Locate the matrix of move cells in
  the image and carve out one image per cell.
\item
  \textbf{OCR} (\cref{sec:arch:ocr}). Run each cell image through the BiLSTM
  recognizer and a CTC beam decoder constrained by a chess-notation
  grammar. This yields, \emph{per cell, a ranked list of candidate moves
  with confidences} rather than a single string. Downstream reconstruction
  is possible only because this full candidate list is carried forward.
\item
  \textbf{Reconstruction} (\cref{sec:recon}). Replay the candidates as a game,
  detecting and repairing the legal-but-wrong reads that cause cascading
  corruption.
\item
  \textbf{Verification}. Surface every flagged move to the operator
  with its ranked candidates; nothing is applied blindly. (During
  interactive replay, only notation-level normalizations (check signs,
  redundant disambiguation, an unmarked capture), forced substitutions,
  and, on dual sheets, a confident and uncontested reading from the other
  sheet are applied automatically; each stays marked in the move list and
  can be reverted. Every other correction needs confirmation, and in batch
  mode every fix is reviewed.)
\item
  \textbf{Export}. Emit standard PGN~\cite{pgnspec} with the seven-tag roster plus
  tournament metadata.
\end{enumerate}

One design intent shapes the whole architecture. The first
three stages (grid detection, OCR, reconstruction) run \emph{ahead} of the
human: OCR and a first reconstruction pass are
computed eagerly, so the operator's time in stage 4 is spent only on the
moves the system could not settle on its own.

\subsection{Grid detection}\label{sec:arch:grid}

Scoresheets are a printed grid of move cells, but a phone photograph of
one is skewed, unevenly lit, and faintly penciled. Robust cell
extraction is therefore a precondition for everything downstream. We
handle it with the \textbf{slide method} (\texttt{grid-slide.js}). Unlike
a global four-corner perspective warp, the slide method never rectifies
the whole page. Instead it aligns to the grid's own structure and extracts each
cell with a small local warp. The pipeline is:

\begin{itemize}
\tightlist
\item
  \textbf{Smart crop.} The grayscale image is thresholded (catching even
  light pencil), its scanner-edge border cleared, and a morphological
  close bridges nearby strokes. A row/column projection then yields the
  content bounding box, trimming surrounding whitespace before any
  structure is inferred.
\item
  \textbf{Connected-component detection and clustering.} Ink components
  are found and clustered into the rows and columns of the grid. The
  number of expected number-anchor columns comes from the user's
  \textbf{sheet profile} (\cref{sec:arch:profiles}). In the common case
  there are two or three per sheet. There are four or six when two sheets
  are photographed side by side and must be split.
\item
  \textbf{Quality-gated merge and header/footer classification.}
  Clusters are scored and merged under a quality gate, and header/footer
  rows (player names, result) are classified out so the move grid is
  isolated. The slide method emits \textbf{no header row}; it starts
  at the first move row.
\item
  \textbf{Hole-based sliding-window alignment.} Each printed move number
  is identified by the topological holes of its digits (in the usual
  printed forms, 0, 4, 6 and 9 have one and 8 has two). For each anchor column, the detector slides the
  expected sequence of move numbers over the detected rows and keeps the
  offset whose hole signature fits best. This assigns a move number to
  every row and separates header and footer rows, even when some printed
  numbers are missing or overwritten. The method is named for this step.
\item
  \textbf{Direct mini-warp extraction.} Each cell is finally extracted
  with its own small perspective correction, as in Majid and Eicher's default pipeline~\cite{majid2022}, rather than one global warp of the page. Local warps absorb the page's curl and skew per cell,
  which a single homography cannot.
\end{itemize}

Real input fails most often at two practical sub-problems.
\textbf{Side-by-side sheets:} when both players'
scoresheets are photographed in a single frame, the image must first be
split in two. The split, not the per-sheet detection, is the
fragile step. Intensity-based seam finding fails on the inputs
that matter: phone photographs with a lighting gradient, faint pencil,
and a thin, skewed gap between the sheets. The robust split instead
finds the \emph{printed grids'} own boundary. It first detects both halves'
move-number columns on the unsplit image and places the seam in the widest
gap between them. If that fails, it tries a seam located from the printed
table rules, then a column-count heuristic, and only last an ink-density
valley. \textbf{Printed vertical rules:} a cell's printed left border
frequently bleeds into the extracted cell image. There the recognizer
can read it as a character stroke, and a handwritten \emph{c4} becomes
\emph{Nc4}. A cleaning pass identifies the rule by a continuity gate.
Only a stroke that runs essentially edge-to-edge through the cell
qualifies; genuine pen strokes start and stop inside the
cell and do not satisfy it. \emph{How} the detected rule is then removed
turned out to matter more than whether. Our first implementation erased
only the rule's own dark pixels, and made matters worse. Erasure
leaves the rule's anti-aliased fringe behind, and a faint, broken
vertical remnant is more dangerous than the intact rule. The recognizer
was trained on cells that include printed borders, so it treats a solid
full-height bar as page furniture and ignores it. The half-erased
remnant is outside its training distribution and is read as a pen
stroke. The result is confident phantom piece letters (\emph{c4} decoded as
\emph{Nc4} at 0.98 confidence, on both players' sheets at once, since
both carry the same printed rule). Replaying the recognizer offline on
one fully verified game (124 cells across both sheets) made the
comparison exact. Pixel erasure yielded 59 correct top reads, and no
cleaning at all yielded 62. \emph{Cropping} the cell at the rule's right
edge, which removes the rule, its fringe, and everything left of it in one
cut, yielded 75. The production system crops. The general lesson is that
partial artifact removal can be worse than none, because it manufactures
inputs the recognizer has never seen. The cleaning remains transient and
is applied at the OCR boundary only. The operator always reviews the raw,
uncleaned ink.

We regard the slide method as a contribution in its own right, though a
narrower one than it first appears. Prior
work already handles phone photographs taken at real events. Majid and
Eicher build a grid-line mask, follow cell borders to recover the cell
contours, and correct the perspective of each box individually, with
manual four-corner marking as an alternative~\cite{majid2022}. Commercial scoresheet
products clearly solve this stage too, with proprietary, unpublished
methods. We add two things that are, to our knowledge, not described
elsewhere: anchoring the grid on the printed move-number columns, and a
sliding-window alignment of the printed move numbers by the holes in their
digits.

For a batch tool, it matters that \textbf{grid detection never blocks the
pipeline}. When a sheet's grid cannot be confidently recovered, the
failure is flagged and processing continues to the next game.

\subsubsection{Sheet profiles}\label{sec:arch:profiles}

A \textbf{sheet profile} declares a scoresheet's layout: column
count, rows per page, header/footer rows, page count, and starting move
number (for continuation sheets that begin partway through a game).
Profiles are plain data. The system ships presets for a range of
clubs and federations and lets users author, export, and import their
own. The profile feeds the grid detector its expected column count and
row structure. With it, one detector can serve sheets from many
different tournaments without per-sheet tuning.

\subsection{The OCR layer}\label{sec:arch:ocr}

\textbf{We build directly on Eicher et al.~\cite{eicher2021,majid2022} for handwriting
recognition.} Their BiLSTM+CTC recognizer, trained on the HCS dataset
(\cref{sec:related:hcs}), is exported to \textbf{ONNX} and run client-side through
\textbf{ONNX Runtime Web}~\cite{onnxruntime}. For each cell, the network emits a sequence
of character logits, which a \textbf{CTC beam decoder} turns into ranked
move candidates. Three design choices around this borrowed core are ours
and matter for reconstruction.

\textbf{Grammar-constrained, dual decoding.} The beam decoder is
constrained by a chess-notation grammar, so the candidate list is drawn
from the space of plausible move strings rather than arbitrary character
sequences. We run the decode \emph{twice} per cell. One pass uses a
\textbf{strict} SAN grammar. The other uses a \textbf{lenient} grammar
that also accepts forms common on real scoresheets but outside SAN:
file-only pawn captures (\texttt{cd} for \texttt{cxd5}), piece-takes-piece
captures (\texttt{BxN}), and long algebraic moves (\texttt{e2-e4}). The
lenient readings are kept as a separate, down-weighted list. In the
fix-finder they never create a candidate on their own, but add evidence to
a legal move it has already proposed; the interactive quick-fix list does
show a lenient reading that normalizes to a legal move. Two further deviations need no grammar,
because the recognizer's character set excludes them. An explicit pawn
letter (\texttt{Pe4}) is outside the vocabulary and is not recovered.
Castling written with zeros (the form used in the FIDE Laws' own
notation appendix~\cite{fide}, as against PGN's letter
\texttt{O-O}~\cite{pgnspec}) is read as the letter.
Some lenient readings name no definite move in isolation. A
shorthand like \texttt{cd} denotes a capture only the position can
disambiguate. Lenient readings are therefore carried as raw text and
normalized to standard notation later, \emph{against the reconstructed
position}, because their meaning is itself a function of reconstruction state.

\textbf{Constrained re-OCR.} Each cell's raw logits are cached
(\texttt{storedLogits}). When reconstruction later flags a cell, the
system can re-score \emph{only the legal moves at that position}
directly against the cached logits, via CTC forced alignment
(\texttt{decodeConstrained}). The operator sees this among the Quick
Fixes as the ``fresh re-OCR of one cell''. It asks the recognizer a
narrower question: ``of the moves that are legal here, which does the ink
best support?''. That question can only be posed once
chess context is available. The same forced alignment also scores
\emph{every} candidate the fix-finder proposes, as one of the two learned
ranking signals of \cref{sec:recon:learned}. For that, the logits must outlive the OCR
session. They are written beside the OCR text as a compact float16
sidecar (about 3.5\,KB per cell), so a tournament reopened days later
ranks exactly as it did on first read. Quantizing to float16 changed no
ranking decision on our corpus.

\textbf{An explicit character-confusion model.} Downstream of the
decoder, every candidate comparison rests on a model of how handwritten
chess characters are actually confused. The model holds graded similarity
scores for 50+ directional character pairs. They are seeded from the
confusion matrix published by Horváth and Dreher~\cite{horvath2020} and
recalibrated empirically against our BiLSTM's behavior. The pairs encode
handwriting geometry rather than typographic resemblance. Truncated
ascenders collapse \texttt{K} into \texttt{R} and \texttt{R} into
\texttt{B}. The piece letter \texttt{B} collides with the file letter
\texttt{b} where it hurts, in capture notation (\texttt{Bxc4} vs
\texttt{bxc4}). The most instructive pair is
\texttt{g}↔\texttt{a}, among the strongest in the table, because
its cause is physical. A handwritten \texttt{g}'s
descender drops below the writing line and out of the cell, so the body
that remains \emph{inside} the cell is an \texttt{a}. The pipeline
exploits this physics directly. When a cell reads \texttt{a}, the top
strip of the cell \emph{below} is searched for an orphaned descender
blob, and finding one boosts the \texttt{g} reading's confidence. Ink
in a neighboring cell thus serves as evidence about this one. The model
supplies the similarity term of candidate scoring (\cref{sec:recon:scoring}) and the
single-character auto-corrections of \cref{sec:recon:fixfinding}.

\textbf{Recognition accuracy and domain shift.} Our retraining of the
architecture\footnote{The trained recognizer is published as an ONNX
model~\cite{zugwisemodel} at \url{https://huggingface.co/GerhardTrippen/chess-ocr-bilstm} and
is fetched directly by the client at runtime.} (from scratch, without IAM
pretraining, with modified preprocessing) reaches \textbf{82.9\% raw
move-recognition accuracy on
held-out HCS writers}. This matches the 82.5\% raw-network figure Eicher et
al.~report for the same architecture~\cite{eicher2021}. The journal version reaches
87.8\% raw (89\% with manual four-corner marking) with IAM pretraining and
more training data~\cite{majid2022}. Their higher headline numbers include
post-processing and legality checking, which in our pipeline belong to
the reconstruction layer. Our regression corpus (\cref{sec:eval:groundtruth}, \cref{sec:eval:ablation} and \cref{sec:eval:priorwork})
consists of real club scoresheets, with phone photography,
varied layouts, and a wide range of handwriting. On it, the raw top read matches
the aligned game record on \textbf{84.0\%} of plies over the six
single-sheet games (288/343; 82.4\% including the two dual-sheet games'
merged reads). The corpus average thus survives the domain shift
essentially intact. What does not survive is the \textbf{per-game
floor}: individual games range from 94.6\% down to \textbf{68.9\%}. Two
caveats apply. The exact-match comparison is alignment-aware with check
symbols stripped. And the corpus was assembled for regression value, so
it is deliberately rich in difficult sheets. The mean therefore likely
understates typical club input, while the spread is an accurate picture.
The reconstruction layer (\cref{sec:recon}) is therefore engineered
for the 69\%-accuracy tail, where roughly one move-box in three arrives
wrong.

\subsubsection{The OCR worker pool}\label{sec:arch:workerpool}

OCR is the pipeline's compute bottleneck, and a deployment constraint
shapes its parallelization. Because the public deployment is
served from \texttt{github.io}, the page is \textbf{not cross-origin
isolated}. \texttt{SharedArrayBuffer} is therefore unavailable, and with it multi-threaded WASM. (A service worker can add the required isolation headers, following the coi-serviceworker pattern, but we have not adopted it.) ONNX Runtime Web runs \textbf{single-threaded}
on one CPU core. We recover parallelism at the process level instead, with a
\textbf{pool of dedicated OCR workers} (\texttt{ocr-pool.js},
\texttt{ocr-worker.js}). The pool is sized to roughly \texttt{min(4,\ cores/2)},
and each worker owns its own ONNX session and runs at most one inference at a
time. Per-cell OCR is dispatched across the pool, giving near-linear
speedup up to the pool size. In practice OCR wall-clock drops roughly 4×
with four workers (an informal measurement, not a controlled
benchmark).

Dispatch uses \textbf{ply affinity} rather than next-idle scheduling: a
cell is routed to \texttt{worker\ =\ ply\ mod\ poolSize}. The choice is
load-bearing. Constrained re-OCR reads the cached
logits \emph{inside} the worker that produced them, so a cell and its
later re-OCR must land on the same worker. In dual-sheet mode both
sheets' copies of a ply therefore share a worker, so a constrained
re-OCR can see both readings. These OCR workers deliberately do
\textbf{not} load the Python runtime. They own only the
ONNX/logit/beam-decode path, which keeps them lightweight.

\subsection{Batch mode: from a tournament folder to a round
PGN}\label{sec:arch:batch}

The use case that motivates the system is a whole tournament. Batch mode is the same pipeline arranged for that
scale, and the single-game workflow is its special case: a game list with
one entry.

\textbf{Inputs.} The operator supplies a folder of scans and, separately,
the tournament's pairing file (SwissManager, SwissSys, or a
chessmanager.com export). The pairing file provides the players, ratings,
boards, and results; the folder provides the images. File names on real
tournament scans are inconsistent, so games are identified primarily from
the directory structure (section, round, board) and only secondarily from
a naming convention. PDFs from scanning apps are rendered to images in
the browser.

\textbf{Queues that run ahead of the operator.} OCR workers process the
sheets in order and write the recognizer's output, the cell geometry, and
the cell logits beside the scans, so a tournament reopened later is never
re-recognized. A reconstruction queue runs greedy (\cref{sec:recon:fixfinding}) on each
game as soon as its sheets are read. A game greedy cannot
finish is escalated to beam and then to Dijkstra (either can be switched
off). When beam's queue is idle, beam also works ahead on the hardest game
greedy has not yet finished, so hard games often have a beam result early. The
operator therefore starts reviewing the first game while later ones are
still being read, and does not wait for OCR.

\textbf{Triage by review effort.} Each game is classed by how much review
it is likely to need: \emph{quick} when the searches that ran agree, every
fix is highly similar to the written text, and there are few fixes;
\emph{standard} for minor disagreement and a moderate number of fixes;
\emph{deep} when the searches diverge, any of them fails, or the fixes are
many. The classes order the operator's work without reducing it.
Agreement among the searches is only evidence, and nothing is applied
without review.

\textbf{Verification replaces ``apply''.} A search result is never
applied in one step. The operator walks through it fix by fix, with the
scoresheet cell, the board, and the ranked alternatives in view, and
there is deliberately no ``confirm all remaining'' control. When the
operator overrides a fix, the searches are re-run from that ply onward
only, so earlier confirmed work stands.

\textbf{Export.} A round is exported as one PGN file with headers filled
in from the pairing file and a source tag identifying the tool, along
with an error report for games that could not be completed. Where the
browser allows, results are written straight into the scan folder;
otherwise they are offered as a single download. The per-game effort
figures of \cref{sec:eval:results} were logged by this workflow.

\subsection{Client-side, offline-capable deployment}\label{sec:arch:deploy}

Zugwise runs entirely in the browser; no scoresheet image, tournament
file, or move ever leaves the device.\footnote{The public deployment
counts anonymous page views through a cookieless analytics service, which
records that the page was opened and nothing about any game. The
application code and the two models are fetched on first load; nothing
about any game is transmitted.} The guarantee is both a privacy one
and an operational one: there is no server to run, and so zero
marginal cost per game. Three pieces make it work:

\begin{itemize}
\tightlist
\item
  \textbf{Reconstruction in the browser via Pyodide.} The reconstruction
  engine (\cref{sec:recon}) is Python, including \texttt{python-chess}~\cite{pythonchess} for move
  generation and legality. It runs in-browser under \textbf{Pyodide}~\cite{pyodide}.
  The reconstruction modules are fetched from \texttt{/backend-python/}
  and executed into a single flat namespace. Because Pyodide has no
  package structure here, each module's local \texttt{import} statements
  are stripped before execution. Fifteen modules load in dependency
  order (\texttt{data\_structures}, \texttt{helpers},
  \texttt{similarity}, \texttt{chess\_quiescence}, \texttt{absurdity},
  \texttt{play}, \texttt{constraints}, \texttt{missing\_moves},
  \texttt{lenient\_normalize}, \texttt{logits\_io}, \texttt{ctc\_align},
  \texttt{fix\_finding},
  \texttt{full\_game\_search}, \texttt{dijkstra\_search},
  \texttt{validation}). The same loader serves both the main
  reconstruction worker (\texttt{zugwise-worker.js}) and the lighter
  background search workers (\texttt{search-worker.js}).
\item
  \textbf{A separated worker layout.} ONNX OCR runs in the worker pool
  (\cref{sec:arch:workerpool}); Pyodide reconstruction runs in \texttt{zugwise-worker.js};
  long-running searches (beam, Dijkstra) run in background search
  workers so they never block the UI or OCR. The OCR workers carry no
  Pyodide; by default the Pyodide workers carry no ONNX. The human-move prior
  (\cref{sec:recon:learned}) is a second ONNX model with a worker of its own. Before
  a search starts, it scores the game's positions in one batched call and
  hands the Python side a plain table of log-probabilities. If the model
  fails to load, the prior drops out of the ranking and the search runs
  as before.
\item
  \textbf{PWA, with a two-repository deploy.} The app is a Progressive
  Web App. A service worker caches the models, code, and assets on first
  load, after which the app runs fully offline; a new version installs in the
  background and takes effect when the user reloads. Development happens in a private repository. A deploy
  script copies the shipping subset (frontend, the fifteen in-browser
  Python modules, piece sets, icons) into a public repository served by GitHub
  Pages at \url{https://gerhardtrippen.github.io/Zugwise/}, with an audit
  step that flags any source file not on the deploy manifest.
\end{itemize}

Running everything locally has an architectural payoff beyond privacy.
It means the system can be put in the hands of a
tournament organizer with no infrastructure, no account, and no per-game
cost. Such a deployment is the precondition for the use case the paper is
ultimately about: digitizing whole tournaments that would otherwise
never be archived at all.

\section{Game Reconstruction}\label{sec:recon}

\subsection{Reconstruction, not document analysis}\label{sec:recon:philosophy}

This work stops treating a scoresheet as a document to be transcribed
and treats it instead as a \emph{game to be reconstructed}. The two
framings diverge where it matters. A document-analysis pipeline asks,
for each cell, ``what character string is written here?'' Having
answered, it checks only that the resulting move is legal. A
reconstruction pipeline asks instead, ``what sequence of moves was
actually played, given this noisy evidence?'' The gap between these
questions separates a system that silently corrupts a game from one
that recovers it.

The corruption is real. When an OCR error produces a move that is
\emph{illegal}, every prior system catches it, because the error is
loud. The dangerous errors are the quiet ones: an OCR misread that
happens to be a \emph{legal} move in the current position. A fully
automated pipeline accepts it and advances the board to the wrong
position. It then interprets every subsequent move against that wrong
position. We call this \textbf{cascading corruption}. This structural
failure mode motivates the entire architecture. Horváth
and Dreher \cite{horvath2020} encountered it independently in 2020. They
attempted iterative game-state tracking and abandoned the approach when
a single wrong move corrupted all positions downstream.

\subsection{Absurdity detection as a quiescence application}\label{sec:recon:absurdity}

The signal that an OCR reading is wrong is rarely visible at the
character level. It shows at the \emph{board} level. A legal-but-wrong
move tends to leave the position in a state that human games rarely
produce. The most characteristic case is a piece left \emph{en prise},
hanging uncontested, while the opponent does nothing about it for move
after move. A hung queen is usually taken at once. A scoresheet in which
a queen sits capturable move after move while both players ignore it is
therefore suspicious, and more often a record of an OCR error than of
human play.

We detect this with a two-layer mechanism that repurposes the quiescence
search of a chess engine.

\textbf{Layer 1 (instant).} After each move we run a cheap, pin-aware
count of attackers and defenders on each of the moving side's pieces. In
roughly 99\% of positions nothing is flagged and we continue
immediately.

\textbf{Layer 2 (quiescence, only when flagged).} When Layer 1 finds an
apparently hanging piece, we ask whether capturing it is
\emph{genuinely} free. A quiescence search resolves the exchange and
reports the net material outcome. The search is minimax with
alpha-beta pruning~\cite{knuth1975} over captures, checks and
promotions (and every legal reply when in check). It runs for the capture
and up to eight further plies and is scored by material at the point where
it stops, with an imminent promotion counted as worth eight points; the
side to move may ``stand pat'', except when in check or facing an
imminent promotion. If capturing the piece loses material for the
capturer within that horizon (a trap, or a pin that wins material
back), it is not treated as absurd and we continue. Layer~2 flags the
piece only when the search finds that capturing it wins at least two
points of material (pawn~1, minor piece~3, rook~5, queen~9). It treats
that as evidence of an upstream OCR error. The evidence is fallible,
since a human can overlook a tactic.

\textbf{The stop rule.} Layer~1 treats a piece as apparently hanging when
it is attacked and undefended, or attacked by a piece of lower value
even if defended; pins are respected on both sides. The replay then
stops at a move in four cases. (i)~\emph{Bad trade}: the move is a piece move to a
square the opponent attacks, or a capture by a more valuable piece, and a
static exchange evaluation (SEE) of the destination square shows the mover
losing at least two points, unless the move was the only legal reply to
check, ends the game, or has the compensation described below. SEE
alternates the cheapest legal captures on that square alone, and each side
is free to stop. (ii)~\emph{Piece left hanging}: after the move, a minor
piece or larger of the side that moved is genuinely hanging by Layer~2, and
the opponent's recorded reply neither captures it nor gives check. A piece
worth no more than what the move captured is exempt. A fallback also flags
an undefended piece whose capture gains less than two points, under further
conditions given in the code. (The interactive validator applies the same
test but exempts only a reply that captures the piece.) (iii)~\emph{Persistent
absurdity}: a minor piece or larger that the move leaves hanging is followed
across both sides' plies until it is captured, moves, or is no longer
hanging. If it is still hanging two plies later with no reaction in the
record, the replay stops at the move that first left it hanging. Only a
free choice counts as no reaction: a ply on which the move gives check,
answers a check, captures at least as much material, or is made by an
attacker that could not legally take the piece does not count, and the rule
never fires when the next recorded move takes the piece. In our recordings
it never fires, because stops (i), (ii) and (iv) catch these positions first.
(iv)~\emph{Ignored free capture with check}: after the move, the opponent
can win material with a capture that gives check (a pawn counts here), and
the recorded reply neither gives check nor captures as much. The replay stops
at the move that allowed it. On the 90 verified games of the two round-robin
sections of \cref{sec:eval:learned} (all 45 games of each, including the second section's
forty-fifth), this rule fires four times on correct moves.
Pieces left hanging while answering a check they could not help with
are exempt, as are exchanges with visible compensation. For a bad trade,
that is a reply, after the opponent's cheapest recapture, that mates in one
or two moves or wins at least the lost material with a free capture; for a
hanging piece, material the move itself captured, an immediate mate, or a
promotion that survives. The
reason for the stop is shown to the operator. The stop is a question,
not a verdict.

In short: \emph{we use the same machinery a chess engine uses, but
instead of evaluating the quality of a position, we test whether a
specific capture is free.} The engine asks
``what should you play?''; we ask ``could a human plausibly have
produced this?'' The full quiescence search is far too slow to run
everywhere. Running it at every position made replay roughly seventy
times slower in our measurements, so the hybrid design pays for it
only in the rare flagged positions.

\subsection{The plausibility boundary: a matched pair}\label{sec:recon:matchedpair}

The hardest cases are those where chess knowledge and reconstruction
fidelity pull in opposite directions. Two games from the same
tournament, in the same rating range, make the tension concrete. Read
together, they show the limits of judging a reading from the position
alone (\cref{fig:matchedpair}).

\begin{figure}[t]
\centering
\begin{minipage}[t]{0.46\textwidth}
\centering
\chessboard[setfen={2krr3/p4p2/2pp4/1p5p/4P3/P2P1Pq1/2Q3P1/R4KR1 b - - 8 29},
            boardfontsize=16pt, showmover=true]
\\[2mm]
{\small\textbf{Game A} — Black to move. The reading \texttt{Rd7} allows
\texttt{Qxc6+}, a free capture \emph{with check} the king cannot recapture;
\texttt{Kc7} defuses it (\texttt{Kxc6} recaptures the queen). Chess
knowledge \emph{resolves} the ambiguity.}
\end{minipage}\hfill
\begin{minipage}[t]{0.46\textwidth}
\centering
\chessboard[setfen={4Qr1k/ppq3pp/3p1b2/3P4/3p2P1/5N1P/PPP2P2/3RR2K b - - 0 22},
            boardfontsize=16pt, showmover=true]
\\[2mm]
{\small\textbf{Game B} — Black to move. \texttt{Rxe8} captures the queen but
allows \texttt{Rxe8\#}; the player did this deliberately and resigned to the
mate. Here the same chess knowledge points the \emph{wrong} way.}
\end{minipage}
\caption{The plausibility boundary (\cref{sec:recon:matchedpair}). Two games from the same tournament
and rating range, opposite lessons: in Game~A a position-level fact
(a free capture with check) disambiguates two legal readings correctly; in
Game~B the analogous reasoning would penalize the move that was actually played.
The information that separates the cases lies in the record, not the position,
which is why the system keeps a human in the loop.}
\label{fig:matchedpair}
\end{figure}

\textbf{Game A: chess knowledge \emph{helps}.} At one point the OCR
offers two legal readings of Black's move: a rook move, \texttt{Rd7},
and a king move, \texttt{Kc7}. The two readings differ in \emph{two}
characters at once. One is the piece letter (the R↔K confusion is among
the most common in handwriting), the other the file letter (d↔c), and
both readings are legal. But after \texttt{Rd7}, White has
\texttt{Qxc6+}, a capture of a pawn that also gives check, and the king
cannot recapture. A free capture with check is an immediately forcing
threat. The player to move must respond to the check on the very next
ply, and they do not get to ignore a free pawn taken with tempo (though
a player can of course overlook the capture before it is played). After
\texttt{Kc7}, by contrast, \texttt{Qxc6+} is no longer free:
\texttt{Kxc6} recaptures the queen, and quiescence detects this. Here
the chess-aware reading resolves the ambiguity correctly. \texttt{Kc7}
is what was played, and the position confirms it.

The \textbf{free-capture-with-check penalty} now handles this case in
code. The detector runs quiescence only on captures that also give
check, so in the overwhelming majority of positions, where no capture
gives check, the cost is effectively zero. It penalizes a candidate move
when it would hand the opponent a quiescence-verified free capture with
check. It mirrors the missed-free-capture penalty, with the same logic
in the opposite direction. The penalty is only evidence. A player
can overlook such a capture, so the penalty lowers a candidate's rank
without excluding it.

\textbf{Game B: chess knowledge \emph{hurts}.} Here the OCR offers a
capture, \texttt{Rxe8}, that is legal but allows the opponent an
immediate mate: after \texttt{Rxe8}, the reply is \texttt{Rxe8\#}. A
naive ``good move'' heuristic would reject \texttt{Rxe8} outright. Why
would anyone walk into mate-in-one? But Black did so, deliberately. The
capture was a signal of resignation, and White mated at once, a
``resignation capture''. (The intent is known from the event itself, not
inferred from the moves; the scoresheets alone show only the sequence.)
Chess quality says reject; the record says accept. A position-only check
does not see the end-of-game context. When a candidate allows an
immediate mate and the scoresheet records that mate as its last move,
the mate is consistent with how the game ended. (An abrupt end of the
sheet alone is weaker evidence, since recording can stop.)
(This penalty-to-bonus ``flip'' is a design insight we describe in \cref{sec:disc:helps};
it is not yet implemented.)

Same tournament, same rating range, opposite lessons. In Game A,
applying chess knowledge gets the reconstruction right. In Game B,
applying the same kind of chess knowledge gets it wrong. Tactical
plausibility is fallible. It supports the correct reading in one case
and would penalize a genuinely played move in the other. The
continuation, the end of the record and the handwriting provide the
additional evidence. For this reason the system keeps a human in the
loop. It surfaces the ambiguity, and the operator,
who sees all three, resolves it.

\subsection{Unified scoring and its design rule}\label{sec:recon:scoring}

Candidates are ranked by a single scoring function. Its design follows one
rule, which puts everything above into architectural form:

\begin{quote}
\textbf{The ranker does not use general judgments of what should be played.
Hand-built chess signals are admitted only if they encode the plausibility
floor and are tactically verified. Learned signals are admitted only as
measurements over a population, under the conditions of
\cref{sec:recon:learned}.} The system asks ``could a human plausibly have
written this?'', not ``what should the player have played?''
\end{quote}

\noindent Reconstruction evidence that is not about chess (similarity to the
written text, reach, recognizer confidence, check symbols) lies outside the
rule's scope, because it concerns the record rather than the position. The
rule is a design choice. Added learned and engine-derived ranking terms are
held to an \emph{admission standard}: at settings fixed in advance, they must
improve ranking on games not used to choose them, and their effect on
correctly recorded blunders is reported. The hand-built baseline predates the
standard and is not claimed to pass it. The unchanged prior meets it on a
section digitized after its weight was fixed (\cref{sec:eval:learned}). CTC, at
its fixed weight, also improved ranking there, but its two-sheet form was
introduced while that section was being digitized, so this is not a fully
independent confirmation of the CTC method. A bounded, searched engine
evaluation helped in cross-validation, but its gain was not confirmed on new
games (\cref{sec:eval:engine}). The
hand-built quality signals the rule removed showed no additional benefit once
the prior was present (below).

The natural objection treats this as discarding evidence. A player is more
likely to have produced some moves than others, the argument runs, so a prior
over moves is legitimate Bayesian evidence about what was written, and
excluding it throws information away. The objection is half right, and the
half that is right is part of the system (\cref{sec:recon:learned}). What it gets
wrong is \emph{which} prior. A hand-built quality signal, such as a bonus for
winning material or a penalty for letting the opponent win material, is a
judgment about the position built from a table of piece values. It describes
what a strong player \emph{should} do. Club players often do not play the move
such a signal prefers, so it says little about what was written, and it
penalizes the truth whenever the player blundered. In our measurements the
quality signals we built were not harmful in aggregate, but they added
nothing (below). A \emph{population} prior is a different object. It is an
estimate, learned from some 120{,}000 games by a model never told what a piece
is worth, of how often players in its training band make each move in
positions like this one. It is fallible evidence of the same kind as OCR
confidence, and it answers the reconstruction question.

This gives a working test for any proposed \emph{learned} signal:
\textbf{is it a judgment of the position, or a measurement of what a
population of players actually did?} Population measurements are admitted,
under the conditions of \cref{sec:recon:learned}; judgments are kept out unless they
meet the admission standard (\cref{sec:eval:engine} reports one such test).
Hand-built chess signals are judgments by construction, so for them the test
is stricter. A well-calibrated hand-built feature could in principle carry
predictive information, and the rule rests on our finding that the quality
features we built added nothing beyond the prior (below). Among the hand-built signals, the only
property with real discriminating power is the \emph{plausibility floor}: a
position so degenerate (a major piece left uncontested over several moves, an
avoidable mate-in-one walked into) that it is likelier to be OCR corruption
than human error. That floor is what our permitted penalties measure. They are
checked by tactical search, so that they approximate ``could a human have
produced this?'' rather than ``would a strong engine?'' The same machinery
(quiescence, exchange evaluation) thus underwrites the absurdity penalty,
which is permitted, and would underwrite a winning-capture bonus, which is
not. The tooling does not separate them; the question each asks does, and for
a hand-built signal the requirement of tactical verification, below, applies
on top.

The \textbf{permitted} signals are all reconstruction signals: character
similarity to the OCR text under the confusion model of \cref{sec:arch:ocr}; reach
(does the game continue after this candidate?); OCR confidence and the
candidate bonus; check-symbol matching against the scoresheet;
lenient-grammar parse; the quiescence-backed \textbf{absurdity penalty}; the
quiescence-backed \textbf{missed-free-capture penalty}; the quiescence-backed
\textbf{free-capture-with-check penalty}, applied to candidates at the stuck
ply; and an exact \textbf{forced-mate penalty} that fires when a candidate
would hand the opponent a mate-in-one, checked on the leading candidates in
the final verification pass. While candidates are scored, the absurdity
penalty uses a fast two-ply exchange check; the full quiescence search runs
for an absurdity at the fix ply itself and in that final pass, which covers
the five leading candidates by hand-built score plus up to five from Phase~2
and five others, before the learned terms of \cref{sec:recon:learned} are added.

The forced-mate penalty belongs on the permitted side for the same reason the
absurdity penalty does: it is a plausibility signal, not a quality signal. It
does not ask whether the move is \emph{good}. It asks whether a human
recording a real over-the-board game would plausibly have produced a position
that gives the opponent an \emph{avoidable} immediate mate. (It does not
check whether play continued; that check is the flip discussed in
\cref{sec:disc:helps}.) A rook left hanging beside a queen is suspicious in the same
way, and so is walking into mate-in-one: each can indicate an OCR artifact.
Unlike the capture penalties, the forced-mate penalty needs no search
horizon, because a mate-in-one is computed exactly. What no computation
establishes is whether allowing it was a transcription error or an
oversight, so it too is fallible evidence. Its main exception is the
\textbf{resignation capture}, where the player really does walk into mate
because the game is ending: the end-of-game case of \cref{sec:recon:matchedpair}. The
signal already carries a partial carve-out. It suppresses the penalty when
\emph{every} legal alternative to the candidate also allows the mate, that
is, when the mate was unavoidable. The full flip from penalty to
\emph{confirmation} when the scoresheet ends at the mate remains future work
(\cref{sec:disc:helps}). The forced-mate penalty is thus both a working plausibility
signal and the clearest implemented instance of the tension this section is
built around.

The \textbf{forbidden} signals are hand-written and disabled in the code.
Four are kept with their weights hard-coded to zero, and their variable
names and \texttt{score\_components} keys remain as labeled placeholders, so
the exclusion is visible wherever someone might be tempted to undo it; the
fifth, the material-loss penalty, was removed, and only a comment giving
its rationale remains. They are
of two kinds. Two are chess-quality signals proper: a winning-capture bonus,
and a material-loss penalty that fired when the opponent could win material
in response. Three are unverified plausibility penalties, cruder versions of
the permitted ones: a missed-capture penalty without quiescence, a
pawn-hanging penalty, and a hanging-piece penalty without quiescence.

Among the hand-built chess signals, which ask the plausibility question, we
draw the line between the permitted and the forbidden lists at
\textbf{tactical verification}. A permitted penalty fires only when a search
confirms, to its horizon, that a capture is free or a mate is forced. That
removes the false alarms of static counting (traps, pins, recaptures),
though not the possibility that a human simply missed the tactic. It is the
difference we intend between ``play the engine's move'' and ``a human would
rarely do this.''

The first concrete case is the \textbf{\texttt{Rxe3}
counterexample}. A low-similarity capture of a queen (similarity ≈ 32\%) once
outranked a high-similarity quiet move (\texttt{Rhg8}, similarity ≈ 86\%),
because a winning-capture bonus rewarded grabbing the queen. But
\texttt{Rxe3} was a trap: it loses to the \texttt{gxh8=Q} promotion. The
quality signal pushed the reconstruction toward the wrong, lower-similarity
move. The second is the \textbf{removal of the material-loss penalty}. That
check penalized a candidate when the \emph{opponent} could win material in
response, a chess-quality question (``could the opponent exploit this
position?'') posing as a reconstruction question. Elsewhere in Game A it
penalized the correct fix \texttt{b3} for a vulnerability the position
already contained, a pin that existed before the candidate move. Removing
it, with the regression suite still passing, put the rule into practice: a
600-rated player can hang a rook. That is bad chess faithfully recorded, and
the reconstruction must accept it.

When a new position seems to call for a quality signal, the first responses
are to tune the weights of existing reconstruction signals, improve reach
depth or absurdity accuracy, add a \emph{measured} population signal under
the conditions of \cref{sec:recon:learned}, or accept that the system surfaces the
ambiguity to the user. A hand-written quality signal should not be added on
the strength of a compelling example, however narrow it is: each of the
forbidden signals was locally reasonable when it was added. A hand-tuned
penalty presented as a prior is still a judgment.

For the record, the forbidden signals turn out to be redundant rather than
harmful once the prior is present. We recomputed four of the five with their
original formulas for every candidate on both sections of
\cref{sec:eval:learned}. Restoring them at their original weights changes top-rank
accuracy by $-0.2$ points on each section; both intervals include zero. A
single weight fitted by leave-one-round-out cross-validation does no better
($-0.2$ and $-0.5$ points). Without the prior they would add about one point: what they encode,
that players rarely leave a piece en prise or ignore a free capture, the
prior has learned from games. On these data they show no additional benefit
once the prior is present, so they stay out. Whether a \emph{searched}
evaluation fares differently is tested in \cref{sec:eval:engine}.

\subsection{Two learned signals: re-reading the ink, and a human-move prior}\label{sec:recon:learned}

Every candidate ranked by the scoring function is a hypothesis about one
move. Two kinds of evidence bear on it: how well the ink supports the
candidate's notation, and how likely a player is to make that move in
this position. The hand-built signals approach the first only through
the decoded string. Character similarity compares the candidate with
what the recognizer \emph{wrote down}, not with what it \emph{saw}.
Nothing addressed the second at all, apart from the plausibility floor.
Learned models fill both gaps, each added to the hand-built score with
an empirically chosen weight. We define the terms in the order the system
computes them. Throughout, the reconstruction state (the current move list
and the alignment of the sheets to it) and the candidate list are fixed.

\textbf{Candidates.} A candidate fix $h$ consists of a target ply and a
proposed move $m_h$ for it; $b_h$ is the position before that ply under the
current move list. The same notation at another ply is a different
candidate, and changing an earlier move changes $b_h$. We write $s_h$ for
the candidate's notation as the scorer receives it, the notation stored
with the candidate, with castling written with the letter O, and
$S_{\text{hand}}(h)$ for the hand-built score of \cref{sec:recon:scoring}.

\textbf{Recognition score of one cell.} Let $\Sigma$ be the recognizer's
27 characters and $\Sigma_{\varnothing} = \Sigma \cup \{\varnothing\}$, with
$\varnothing$ the CTC blank. For a cell image $x$ the recognizer emits
$T(x) = 63$ frames, and $p_t(a \mid x)$, for $a \in \Sigma_{\varnothing}$, is
the softmax of frame $t$'s outputs. The system stores the natural
logarithms $\log p_t(a \mid x)$; the code calls these stored values
logits. The map $\mathcal{B}$ turns a path $\pi \in \Sigma_{\varnothing}^{T}$ into
a string by first merging consecutive repeats and then removing blanks, so
$\mathcal{B}(\texttt{N}, \texttt{N}, \varnothing) = \texttt{N}$ and
$\mathcal{B}(\texttt{N}, \varnothing, \texttt{N}) = \texttt{NN}$. With
$\mathcal{A}(s, T) = \{\pi \in \Sigma_{\varnothing}^{T} : \mathcal{B}(\pi) = s\}$,
the score of notation $s$ is its best (Viterbi) alignment,
\[
\ell_V(s; x) \;=\; \max_{\pi \in \mathcal{A}(s, T(x))} \; \sum_{t=1}^{T(x)} \log p_t(\pi_t \mid x),
\]
and $\ell_V(s; x) = -\infty$ if $s$ contains a character outside $\Sigma$ or
$\mathcal{A}(s, T(x))$ is empty. It is a lower bound on the CTC log
probability of $s$, which sums over all paths in $\mathcal{A}$.

\textbf{Both sheets, and notation with and without a check sign.} Let
$x_{h,j}$ be the cell image that sheet $j$ holds for the candidate's ply
under the current alignment, $J(h)$ the set of sheets whose cell for that
ply has stored log probabilities, and $\mathcal{V}(h)$ the set containing $s_h$
and, if $s_h$ ends in a check or mate sign, $s_h$ without it. The same
variants are tried on every sheet. Then
\[
\ell_{\text{OCR}}(h) \;=\;
\begin{cases}
\displaystyle \max_{j \in J(h)} \, \max_{s \in \mathcal{V}(h)} \, \ell_V(s; x_{h,j})
& \text{if this maximum is finite,}\\[4pt]
c & \text{otherwise,}
\end{cases}
\]
where $c = -19.27$ is the mean of $\ell_V$ over all 15{,}195 scored
candidates of an August 2026 recording of the adjudicated section. With the convention
$\max \varnothing = -\infty$, the maximum is $-\infty$ exactly when no sheet
and variant yields a finite score. The fallback applies only after the
maximum: a sheet without stored output is left out of it, and a finite score
below $c$ is kept.

\textbf{Prior.} Let $\mathcal{L}(b)$ be the complete legal moves of position
$b$, with each promotion choice a separate move, $\kappa_b(m)$ the index of
move $m$ among the network's 4{,}096 from--to outputs (squares oriented to
the side to move), and $z_k(b)$ the network's output for index $k$. Then
\[
q(m \mid b) \;=\; \frac{\exp z_{\kappa_b(m)}(b)}{\sum_{u \in \mathcal{L}(b)} \exp z_{\kappa_b(u)}(b)}.
\]
The promotion piece is not encoded, so promotions to the same square share
one index; each still contributes its own term to the denominator and
receives the same probability. With $a = \log 10^{-4}$,
\[
r_{\text{prior}}(h) \;=\;
\begin{cases}
\log q(m_h \mid b_h) & \text{if the prior has scored } h,\\
a & \text{otherwise.}
\end{cases}
\]
The prior has scored $h$ if $b_h$ can be rebuilt, $m_h$ parses there, and the
score is in the table the search uses (\cref{sec:recon:fixfinding}).

\textbf{Combined score.} The ranking score is
\[
S(h) \;=\; S_{\text{hand}}(h)
\;+\; \lambda_{\text{OCR}}\,\bigl(\ell_{\text{OCR}}(h) - c\bigr)
\;+\; \lambda_{\text{prior}}\,\bigl(r_{\text{prior}}(h) - a\bigr),
\]
with $\lambda_{\text{OCR}} = 3$ and $\lambda_{\text{prior}} = 10$, each chosen by
cross-validation (\cref{sec:eval:learned}). Candidates are ranked by
decreasing $S$, and ties keep the order in which the candidates were
constructed. A candidate the recognizer or the prior cannot score receives
exactly zero from that term. For CTC that zero is neutral, since $c$ is a
typical score. For the prior it is a floor: an unscorable candidate ranks as
if $q = 10^{-4}$, 39 points behind a move with $q = 0.005$ and 85 behind one
with $q = 0.5$. This matters most for beam and Dijkstra, which use the table
scored at launch (\cref{sec:recon:fixfinding}). For example, a best alignment score of $-15$
contributes $3(-15 + 19.27) = 12.81$ points, and one of $-25$ contributes
$-17.19$. Taking the better sheet is a ranking choice, not a joint
probability of both sheets, and we do not interpret $S$ as a calibrated
posterior. The prior has no position-dependent reference, since $a$ is one
global constant. The candidates in one list can come from different plies,
so a move in a position with few legal moves scores higher than an equally
typical move in a position with many. A per-position reference would
remove that bias toward forcing positions; we have not measured its effect.

\textbf{Re-reading the ink (CTC re-scoring).} The recognizer is trained
with connectionist temporal classification (CTC)~\cite{graves2006}, which
defines a probability of a transcription \emph{given} the image. The
best-alignment score $\ell_V$ is the quantity the browser decoder
computes. The score never compares strings. It therefore still works
where string similarity is uninformative, for example when the decoder's
top read is a legal but wrong move that shares few characters with the
correct one. The term is applied after the candidate lists from every
phase of the fix-finder have been merged. A term added inside the main
formula would have left candidates from the other construction sites an
implicit zero, which for a negative log-probability is a large reward.
The term is measured against a fixed reference (the typical score of a
candidate over the corpus) rather than against the best candidate of
the current list. The same candidate thus carries the same score in
every list the operator or a search algorithm sees. A check or mate sign
the player left out costs nothing, because the notation is scored with and
without it.

\textbf{A human-move prior.} The second signal comes from a policy
network trained, in the manner of Maia~\cite{mcilroyyoung2020maia}, to predict the
move a human player made. Ours is a residual convolutional network of
eight blocks and 128 filters (2.4M parameters, 9.6\,MB as ONNX). Its
input is the current position alone. In a small probe of Maia on our
corpus, adding the move history changed the top move in none of 42
positions, so we dropped history planes (a probe of the top move, not of
the full candidate probabilities the ranking uses). The output is a
logit for each of the 4{,}096 from--to square pairs, renormalized over
the legal moves of the position. Promotions to different pieces on the
same square share one output and therefore one score. The network was
trained on 120{,}000 games from the lichess open
database~\cite{lichessdb} (CC0) in which both players are rated
1900--2300 on lichess, about 8.35 million positions. On positions from
held-out games, it predicts the from--to squares of the move actually
played 48.4\% of the time (its top output over all 4{,}096). The
held-out positions are the final 5\% of the data in game order, so at most
the one game at the cut contributes to both sides. The model is
published~\cite{zugwiseprior} and runs in the browser (\cref{sec:arch:deploy}). We
trained our own model instead of reusing Maia's weights, so that the training
population and the model are under our control. The debt to Maia is the
idea, and it is a large one. Eicher et al.\ likewise suggested using the
most-played moves in a position~\cite{eicher2021,majid2022}.

The prior meets four conditions, and we would require the same of any
future learned signal:

\begin{enumerate}
\def\labelenumi{\arabic{enumi}.}
\tightlist
\item
  \textbf{It is trained on human play, not strong play}, in a band we
  chose to approximate club players. Lichess ratings are not directly
  comparable with over-the-board ratings, so that match is an assumption
  that the three sections of \cref{sec:eval:learned} test. A
  strong-play policy would reintroduce the population mismatch that made
  the quality signals wrong.
\item
  \textbf{It is added, never substituted.} On its own the prior ranks
  the correct fix first less often than the hand-built score does
  (43.2\% against 58.8\%, \cref{sec:eval:learned}). The prior is
  complementary evidence.
\item
  \textbf{Its weight is chosen by cross-validation}, leave one
  tournament round out, not by a compelling example. Every forbidden
  signal arrived with a compelling example.
\item
  \textbf{It does not replace absurdity detection.} In ranking alone,
  the prior absorbs much of the absurdity \emph{penalty}, but not all of
  it: removing the penalty costs 51 top-ranked fixes without the prior and
  20 with it at its shipped weight. Moreover,
  ranking was never the detector's main job. Its main job is to
  \emph{stop} the replay at a legal-but-wrong move, which a ranking
  signal cannot do. Without absurdity detection, most such errors surface
  only plies later, far from their cause, and some are never flagged at
  all (\cref{sec:eval:ablation}).
\end{enumerate}

The prior has an unavoidable downside, with the shape of the failure
that got the quality signals removed. Sometimes a player makes a move
that few players make, and the prior then pushes a more popular move
above the one actually written. On the adjudicated corpus, adding the
prior on top of CTC re-scoring makes the correct fix top-ranked at 152
decisions and costs it the top rank at 48. We accept that trade. Most
displaced fixes stay near the top. Of the 48, 35 fall to second place and 46 remain in the top three, still in
front of the operator.

One further observation bears on how such a prior should be built. A
variant trained only on games with at least ten minutes of base time
(closer to over-the-board play) predicted held-out moves slightly
\emph{better} and made no significant difference to the ranking (\cref{sec:eval:learned}). How well a
prior predicts moves is not a reliable guide to how much it helps
reconstruction. Only the reconstruction measurement is.

\subsection{The fix-finding pipeline}\label{sec:recon:fixfinding}

When reconstruction stops, the fix-finder engages in phases of
increasing cost and widening scope. A stop occurs at an illegal move,
at a position the absurdity detector of \cref{sec:recon:absurdity}
refuses to play through, and at the review stops: where the two sheets'
best legal readings differ, where another sheet names a different piece
(\cref{sec:recon:dualsheet}), and at an ignored free capture with check (below). The phases are more than a performance ladder.
Each widens the \emph{hypothesis space} about where the true error
lies, from ``the stuck move itself is misread'' outward to ``an error
several moves ago, before anything looked wrong.'' Throughout, the
search proceeds one ply at a time and streams its candidates as it
finds them. The operator sees suggestions immediately and can act or
interrupt at any point, since the pipeline is built to feed a human
decision. Every substitution candidate, in every phase, is scored by the
unified function of \cref{sec:recon:scoring}; insertion candidates, described
below, are ranked separately.

This subsection reports the pipeline's \emph{decision rules}: its
triggers, windows, and thresholds in plies and similarity. The point
coefficients of the hand-built signals (apart from the Dijkstra
constants given below), their gating conditions and the
missing-evidence conventions are specified in the released code
(\texttt{fix\_finding.py},
the \texttt{unified\_score} construction and its \texttt{score\_components},
at the commit given under \emph{Code, models, and data}). The learned
terms use $\lambda_{\text{OCR}} = 3$ with reference $c = -19.27$ (a candidate
without logits receives no term) and $\lambda_{\text{prior}} = 10$, with
an unscorable candidate given $\log 10^{-4}$. An unscorable candidate is
one whose position the browser has not scored, or whose move cannot be
matched in it. For greedy, the table is refreshed after each search step
for the new move line, up to 40 times per search; beam and Dijkstra use the
table scored at launch. We have not measured how often a candidate falls
outside it. The hand-built coefficients were hand-tuned against a small
adjudicated regression corpus, and they leave little on the table. On
the corpus of \cref{sec:eval:learned}, the best linear
re-weighting of the hand-built signals that we tested, fitted
leave-one-round-out, gains 3.1 percentage points of top-rank accuracy. Each
learned signal of \cref{sec:recon:learned} gains more on its own (\cref{sec:eval:learned}). On this
corpus, new evidence improved ranking more than re-weighting the existing
signals did.

\textbf{Anchors and boundaries.} The search is anchored at the
\textbf{stuck point}: the illegal ply, or, when the stop came from
absurdity detection, the ply at which the implausibility appears
(square-level heuristics below still reason about the move the
scoresheet wanted to play). The other boundary is the \textbf{validation
frontier}, the last ply the operator has explicitly confirmed.
Confirmed plies bound the \emph{default} search window. Plies the
operator has \emph{locked} as definitely correct are excluded from
modification unconditionally, in every phase. The frontier is not taken on
trust, however. Before each search the validator re-checks the confirmed
prefix. It stops at the first absurdity, sheet disagreement, or other
forced stop that the operator has not approved, so a confirmation cannot
carry the frontier past a ply nobody reviewed.

\textbf{Quick Fixes (instant).} Before any search, the system presents
\emph{all} legal OCR alternatives at the stuck point, unfiltered by
confidence threshold. The operator sees these first, and they resolve a
large fraction of stuck points with no search at all.

\textbf{Phase 1: frontier to stuck point.} Phase 1 examines every ply
from the validation frontier to the stuck point, \emph{most recent
first}. The reversed order pays off twice. First, the most common case
(a misread at the symptom ply itself) is found immediately. Second, it
enables an early exit. Suppose a candidate completes the entire
remaining game, matches the written text with very high character
similarity (≥ 90\%), and introduces no absurdity. The search then stops,
because such a candidate is almost certainly the correction and
evaluating further plies only delays the operator. One guard protects
the early exit. A candidate identical to the recorded move (``keep as
written'') can never trigger it. The game demonstrably did not play
through as written, and apparent completion under the fast replay can
mask a slow-burning absurdity upstream.

\textbf{Evidence-driven window extension.} The Phase 1 window is not a
fixed radius. The recorded notation carries constraints beyond
move-by-move legality. Violations of those constraints act as beacons
marking \emph{where} the root cause can hide, and each one adds specific
plies to the search. These can include plies \emph{before} the
validation frontier (the beacons are the evidence that justifies
re-opening confirmed moves), though never beyond the stuck point. The
principal beacons are:

\begin{itemize}
\tightlist
\item
  \textbf{Phantom check.} A move recorded with a check symbol that gives
  no check in the reconstructed position is a contradiction between the
  scoresheet and the reconstruction. A shortcut is tried first: if a
  high-similarity move by the same piece \emph{does} give check from
  that position (a one-character slip such as \texttt{Qg5+} for
  \texttt{Qg3+}), only the mismatch ply is added. Failing that, every
  ply up to the mismatch enters the search. The position itself may
  be wrong, and the cause can be anywhere before the symbol.
\item
  \textbf{Duplicate pawn move.} A pawn never returns to a square, so one
  pawn cannot make the same move twice. The same move \emph{text} can
  legitimately recur when a pawn of the same color has at some point
  captured onto that file. After \texttt{c4}, \ldots\texttt{dxc4} and
  later \ldots\texttt{c3} \texttt{bxc3}, a second pawn stands on the
  c-file and can play \texttt{c4} again. Doubled pawns created before
  the first \texttt{c4} allow the same. The detector therefore treats a
  repeat as suspicious only when the preceding record contains no such
  capture, and then both occurrences enter the search. More generally,
  when the stuck ply is one of a duplicated pair, its partner ply is
  targeted directly rather than triggering a full backward sweep.
\item
  \textbf{Blocked path and occupied destination.} When the stuck move is
  illegal because a friendly piece blocks its path or sits on its
  destination square, the specific earlier plies that placed (or failed
  to move) that piece are identified and added.
\item
  \textbf{Blocked piece with a recorded future.} When the stuck move's
  piece \emph{type} can make no legal move at all (for a bishop, no
  legal move on its square color) and yet the scoresheet goes on to
  record further high-confidence moves by that very piece, the piece is
  evidently in play. The reconstructed position must then be wrong,
  because an earlier misread has entombed the piece. The system
  enumerates earlier same-color replacements under which the stuck move
  becomes legal end-to-end and adds those plies to the search. Alone in
  the list, this beacon reasons from evidence \emph{ahead of} the stuck
  point: the remainder of the scoresheet testifies that the piece
  escapes.
\item
  \textbf{Low-confidence readings.} Plies whose top OCR confidence falls
  below 0.80 are marked suspect within the active window.
\end{itemize}

The list is representative rather than exhaustive. The same idea, the
scoresheet's own later record as evidence about an earlier ply, recurs
in smaller structural heuristics that act on scoring rather than on the
search window. A candidate that moves a piece to a square on which the
scoresheet later records the opponent capturing gains a small bonus,
since the recorded capture needs something there to take. A candidate
whose piece the scoresheet later moves to the same square again is
mildly penalized. A candidate that clears the stuck move's destination
square, removes a spurious check symbol, or supplies a missing
disambiguator is preferred, as is, other things equal, a fix nearer the
stuck point. Each asks what the rest of the record implies about the
move in question; none asks whether the move is good.

\textbf{Phase 2: questioning the accepted past.} The true error often
lies before the validation frontier, in an earlier misread that
happened to be legal. Reconstruction and operator alike accepted it,
and it surfaced only plies later. Phase 2 therefore runs whenever any
plies precede the frontier, not only when Phase 1 comes up empty,
because a confirmed prefix can still be wrong. It re-anchors just before the
frontier and searches a deliberately narrow backward window (five plies
by default; configurable up to the whole game), skipping plies Phase 1's
beacons already covered. It may question earlier \emph{auto-applied}
fixes, but never operator locks, and it does not recurse into a Phase 2
of its own.

Phase 2 candidates are then disciplined by the pipeline's most important
scoring rule, the \textbf{advance-past-the-stuck-point requirement}.
Because Phase 2 evaluates candidates against an earlier anchor, their
raw reach is inflated: a pre-frontier change can ``improve'' the replay
while doing nothing about the symptom that triggered the search. After
the phases merge, any pre-frontier candidate whose replay fails to
advance beyond the \emph{original} stuck point is heavily penalized:
70 points if it merely re-reaches the stuck point, and 10 more for every ply
it falls short. A candidate that re-reaches the stuck point and enables a
check the scoresheet records instead loses its reach bonus and half the
flat penalty. The rationale is structural. A
genuine root-cause fix must \emph{explain the symptom away}, whereas a
change that re-arrives at the same stuck position has patched the path,
not the cause. The same discipline governs ``keep as written''
candidates from Phase 2. Keeping an earlier move changes nothing and
cannot unstick the game, so it is dropped from the results. A keep is offered only at the
stuck ply itself, where accepting a flagged move as written is a real
decision.

\textbf{Phase 3: inverting the check geometry.} A targeted phase
handles one recurring trap the first two cannot. Suppose the system
itself added a check symbol during reconstruction (the OCR text had
none, but the move gives check on the reconstructed board). Suppose
also that the very next move is illegal because it fails to address
that check. Two explanations compete. Either the next move is misread
(Phase 1's case), or \emph{the check should never have existed},
because an earlier misread removed the piece or vacated the square that
would have blocked it. Phase 3 pursues the second explanation
geometrically. It computes the squares that would neutralize the check:
the ray between a sliding checker and the king, plus the checker's own
square (for a knight or pawn check, the checker's square alone). It then
searches all prior plies for candidate moves landing on those squares,
scored normally. Rather than widening the window blindly, the check's
geometry dictates which squares an explanatory move must touch.

\textbf{Insertion, not only substitution.} One error class in real
scoresheets is not a misreading at all: the player never wrote the ply.
The omission shifts every subsequent move onto the wrong side to move,
so the record usually turns illegal within a few plies. At that stuck
point, however, a substitution search is looking for the wrong
\emph{kind} of repair. The fix-finder therefore also attempts
\textbf{insertion}. Each legal move at the stuck position is
hypothesized as the forgotten ply, in two forms. (The omission itself can lie a few plies earlier, with
the intervening plies happening to stay legal; insertion is attempted
only at the stuck position, so such a case is recovered by the other
sheet or by the operator.) In the simple form, the recorded stuck move becomes
legal as written once the insertion is made. In the compound form, the
insertion is paired with a similarity-gated correction of the recorded
move. That covers a forgotten ply \emph{and} a misread neighbor at
once, a combination that does occur in practice. Insertion candidates
must advance the replay past everything they touch. They are audited
for absurdities when they complete the game, and they are presented to
the operator explicitly as insertions rather than replacements.
Insertion is the single-record counterpart of the dual-sheet alignment
of \cref{sec:recon:dualsheet}. With two sheets, an omission is recovered
structurally, because the other player usually did write the move; with
one sheet, it must be hypothesized.

\textbf{Reach, asymmetrically.} Across all phases, the workhorse signal
is \textbf{reach}: substitute the candidate, replay the recorded
continuation, and count how many further plies parse legally. Reach is
powerful evidence, since the right correction tends to make the
player's own subsequent moves work again. Reach is nonetheless handled
asymmetrically, by design. Its reward is capped after about five plies
(with only a small per-ply tiebreak beyond). Beyond that point
additional legal plies add little information, and a long-reach capture
could otherwise outrank a high-similarity quiet move. Completion is the
exception. A candidate after which the \emph{entire remaining game}
replays is the strongest reconstruction evidence available, and it
alone retains its full bonuses even when found upstream. An upstream
candidate that merely extends reach may be coincidentally legal, but
one that makes the whole recorded continuation play out has explained
the record. In the other direction, a candidate that advances the
replay not at all is penalized on a sliding scale. The penalty is
\emph{mild} when its similarity to the written text is high (the
operator should still see it; the scoresheet may record a blunder the
replay cannot survive) and severe when similarity is low. An additional
penalty grows the farther such a no-progress candidate sits from the
stuck point. A direct fix at the stuck point that makes progress
receives a modest preference. When the symptom-local explanation works,
it is usually the right one.

Automatic correction during interactive replay is opt-in and off by
default; when it is switched on, it is governed by the
\textbf{``One or Nothing'' rule}. Its candidate space is a fixed family of
repairs to an illegal reading with at most two character changes: inserting
a disambiguator, reading a clipped \emph{g} for an \emph{a}, adding or
dropping the capture mark, swapping a commonly confused piece letter (such
as R↔K), swapping a piece letter for a file letter in a capture (such as
B↔b), and combined piece and square repairs. The repair types are tried in
this order, and a type yields a fix only if exactly one of its repairs is
legal. Uniqueness is checked within each type, so a unique repair of an
earlier type is applied even if a later type also has a legal repair. Zero legal corrections falls back to deep search. Two or more
\emph{always} require user confirmation, because picking the fewest-changes
candidate among several is how silent errors are introduced. Even a unique
fix goes to the operator when it changes the piece letter or drops a
capture, check or mate sign. Within a type, the system never auto-applies a choice
among several candidates (Quick Fixes show them all, for the operator to
choose).

Three search algorithms serve distinct roles. The question we ask of
them is ``what role does each serve in reducing human effort?'', not
``which is best'':

\begin{itemize}
\tightlist
\item
  \textbf{Greedy}: the production workhorse for batch reconstruction.
  Fast enough for the interactive, human-assisted workflow.
\item
  \textbf{Beam}~\cite{lowerre1976}: a second opinion that explores
  multiple reconstruction paths simultaneously. Its width is bounded, so
  it may prune the right path at the boundary. Where beam and greedy
  diverge, there is genuine ambiguity worth human scrutiny; the
  agreement/disagreement is itself a confidence signal. Slower than
  greedy.
\item
  \textbf{Dijkstra}~\cite{dijkstra1959}: a best-first search that is
  cost-optimal under its path cost only when run to completion, without
  the budget and queue limits below.
  The path cost charges each fix a fixed intervention cost (21 points)
  plus its \emph{regret}, the score gap to the best fix proposed at that
  stop: for a path that chooses $h_k$ from the candidate list $\mathcal{H}_k$
  at its $k$-th stop, the cost is $\sum_k \bigl(21 + \max_{h \in \mathcal{H}_k} S(h)
  - S(h_k)\bigr)$. A search state is a whole move list, not a board position,
  and a path counts as complete when it is taken from the queue as the
  cheapest state, not when it is first generated. A path is thus cheap when it makes few interventions and each is
  the locally preferred reading. Alternatives with regret above 42 are
  not explored (both constants were set in February 2026, before the
  learned signals were added, and have not been re-tuned). That
  threshold restricts the search graph. Separately, the deployed search
  runs under a step budget and a queue of at most 50 paths, so in
  practice it is a heuristic like the other two.
  Valuable for understanding the problem space and for checking that
  greedy and beam are not missing fundamentally better solutions. Its
  optimality is bounded in any case. It holds only \emph{within the
  hypotheses the system generates}: the recognizer's candidates, the
  constrained re-reading over legal moves, and the replacement
  hypotheses of this section. The searches replace moves only
  (insertions are offered to the operator at stops), and a Dijkstra path
  holds at most 15 fixes. A correct move that none of these admits
  cannot be found by any of the searches. The stop is still flagged, but
  the operator has to supply the move.
\end{itemize}

In batch mode the three run as an escalation chain: a game that greedy
cannot finish is handed to beam, and then to Dijkstra (each can be
switched off). They are heuristics with different characteristics, not
a hierarchy of trust. Agreement across algorithms is a stronger signal
than any single algorithm's output, and every result is still reviewed
by the operator.

All three share the infrastructure that makes any of them feasible
client-side. Replay results are memoized in per-path prefix caches:
the legality of a move prefix, a path's early-absurdity state, and the
absurdities found along it. Exploring thousands of candidate
continuations therefore does not cost thousands of full-game replays.
Beam and Dijkstra are practical in a browser tab only because of this
reuse.

\subsection{Dual-scoresheet alignment}\label{sec:recon:dualsheet}

When both players' scoresheets are available, the second sheet is not a
separate algorithm but an additional evidence source wired into the same
engine. We align the two move sequences with an adaptation of
\textbf{Needleman--Wunsch}, replacing the amino-acid substitution matrix
with an OCR candidate-pool overlap score. For cell $u$ of the first sheet
and cell $v$ of the second (at this stage not yet known to record the same
ply), let $C_{1,u}$ and $C_{2,v}$ be their candidate readings, compared after
removing check signs and writing castling with the letter O, and
$w_{1,u}(s)$, $w_{2,v}(s)$ their confidences: the decoder's top five beam
probabilities, renormalized to sum to one. The match score is
\[
M(u, v) \;=\; \sum_{s \in C_{1,u} \cap C_{2,v}} \bigl(w_{1,u}(s) + w_{2,v}(s)\bigr),
\]
so $0 \le M(u, v) \le 2$, and a gap costs 0.3. $M$ measures shared candidate
support, not agreement on the preferred reading. In practice it is largely
\textbf{bimodal}: roughly 0.9--2.0 for two readings of the same move, and
near 0.0 for different moves, so most alignment decisions are clear-cut. A
sliding window of $M$ along the plies finds the first place where the
sheets fall out of step; a duplicated row near it is tried first as the
explanation, and otherwise the alignment runs on a short slice around it
and proposes the first gap bracketed by strong matches on both sides
(thresholds and tie-breaking are given in the released code). Each proposal
is an edit, a gap for a missing move or the removal of a duplicated row,
that the operator confirms; automatic application exists but is off by
default. In this way the alignment
handles shift errors (one player's move numbers offset from
the other's), missing moves, and duplicated rows.

The merged result is classified into three tiers. In \textbf{Tier 1}, the
two sheets' best \emph{legal} readings, compared on the board, are the same
move; the ply is locked for the searches by default. In \textbf{Tier 2}, the
chosen reading is legal but only one sheet has a legal reading at all
(including plies only one sheet records). In \textbf{Tier 3}, no reading is
legal, or the two sheets' best legal readings differ; every ply after the
first Tier~3 ply is also Tier~3. Any such difference is a stop for the operator, whatever the
confidences, because here a silent automated choice would risk a quiet
error. At that stop the system preselects the reading $s$ that maximizes
$G(s) = \sqrt{\tilde w_1(s) \, \tilde w_2(s)}$, where $\tilde w_j(s)$ is sheet
$j$'s confidence in $s$, or $10^{-4}$ if sheet $j$ does not list $s$. A
reading that is an immediate bad trade is not eligible when another is.
The three uses of the second sheet differ by design: alignment rewards
shared candidate support, preselection rewards support from both sheets,
and CTC re-scoring (\cref{sec:recon:learned}) lets a legible reading on either
sheet support a candidate.
Tier 2 is the same path we criticize in prior work: a legal-but-wrong reading
on one sheet can win against an illegal reading on the other. Two safeguards
apply. Tier 2 plies are not locked, so they still pass through absurdity
detection, reach and backtracking, and the fix-finder can revise them. And
when the sheet without a legal reading names a different piece (not a pawn)
in its two most confident readings, and that piece can legally reach the same
square, the replay stops and offers that move. Locking is not absolute either:
a Tier 1 ply flagged by absurdity detection is unlocked for the searches, so
an error both sheets share can still be repaired. CTC re-scoring
(\cref{sec:recon:learned}) scores each candidate against both sheets' cells and keeps the
better match.

\section{Evaluation}\label{sec:eval}

\subsection{What to measure: game-level correctness, not
move-recognition accuracy}\label{sec:eval:measure}

Prior work evaluates scoresheet digitization by \textbf{move recognition
accuracy (MRA)}, the fraction of move boxes whose text is recognized
correctly~\cite{eicher2021,majid2022}. For the recognition problem, MRA is the right
measure, and prior work has driven it impressively close to its ceiling.
It answers a different question, however, from the one an archivist
ultimately needs answered. The reason follows directly from
\cref{sec:recon:philosophy}: the errors that matter are not distributed uniformly over
move boxes. A misread that produces an \emph{illegal} move is loud.
Every system since the earliest legality checkers catches it, and it
costs only the effort of a correction. A misread that produces a
\emph{legal} move is silent. A legality-only pipeline accepts it,
advances the board to a wrong position, and interprets every subsequent
move against that wrong position. Under MRA these two errors count the
same, one wrong box each. Under any measure of whether the archived
game is the game that was played, they could hardly be more different.

One consequence is easy to miss. In a pipeline that corrects
illegal readings and accepts legal ones, post-processing removes the loud
errors it can see. The errors that survive are therefore disproportionately
legal-but-wrong readings that no legality check can reach, and, when they are confident, no confidence check either. The
more such a pipeline relies on legality to catch errors, the more of
what remains belongs to a layer above recognition. In this sense the present work
extends prior recognition work and is enabled by its success. Because Eicher et al.~brought recognition this
far, the game-level question becomes the natural next one to ask.

Consider what even an excellent MRA implies at the game level. At 99\%
per-move accuracy, an 80-ply game expects 0.8 erroneous moves, a
figure that requires no assumption about how errors are distributed. If
errors were independent across move boxes, over half of games would
contain at least one (\(1 - 0.99^{80} \approx 0.55\)). In practice
errors are not independent. They cluster by scribe, handwriting
fatigue, and game phase, which concentrates the damage in fewer games
carrying more errors each while leaving the expected count untouched.
Without an assumption about that dependence, per-move accuracy alone does not fix the share of games affected. With 80 plies and 0.8 expected errors, it could lie anywhere between 1\% and 80\%, so game-level correctness has to be measured directly. Whatever the share, the silent, legal-but-wrong errors among them corrupt a game with no flag indicating which games to distrust. The per-box metric saturates just where the per-game
problem begins.

We therefore evaluate on two axes:

\begin{enumerate}
\def\labelenumi{\arabic{enumi}.}
\tightlist
\item
  \textbf{Game-level reconstruction correctness}: does the final
  record agree, move-for-move, with what the players actually played, as
  adjudicated against the original scoresheets?
\item
  \textbf{Human effort}: how much operator attention does each
  archived game cost?
\end{enumerate}

Both axes are measured against the realistic alternative: a competent
human transcribing the same scoresheets by hand. The rest of this section
makes that comparison.

Several corpora and several versions of the system appear below, and it
matters which result comes from which. \Cref{tab:provenance} lists them. In
particular, the expert comparison was carried out \emph{before} either
learned signal existed, and the oracle harness (\cref{sec:eval:groundtruth}) ranks without
the human-move prior, which is applied afterwards to the recorded
candidates.

\begin{longtable}[]{@{}
  >{\raggedright\arraybackslash}p{(\columnwidth - 6\tabcolsep) * \real{0.24}}
  >{\raggedright\arraybackslash}p{(\columnwidth - 6\tabcolsep) * \real{0.22}}
  >{\raggedright\arraybackslash}p{(\columnwidth - 6\tabcolsep) * \real{0.32}}
  >{\raggedright\arraybackslash}p{(\columnwidth - 6\tabcolsep) * \real{0.22}}@{}}
\caption{Which corpus and which system version each result comes from.}\label{tab:provenance}\\
\toprule
Corpus & Used for & System and signals & Reference \\
\midrule
\endfirsthead
\toprule
Corpus & Used for & System and signals & Reference \\
\midrule
\endhead
HCS, held-out writers & Recognizer accuracy (\cref{sec:arch:ocr}) & Our BiLSTM retraining; writer-disjoint split & HCS labels \\
Regression corpus, 8 games (6 single-sheet, 2 dual) & Development, raw recognition accuracy, rank metrics, ablation, baseline (\cref{sec:arch:ocr}, \cref{sec:eval:groundtruth}, \cref{sec:eval:ablation}, \cref{sec:eval:priorwork}) & Release v0.12.40; oracle harness; no cell logits and no prior & Adjudicated game records \\
Crown section, 45 games & Expert comparison (\cref{sec:eval:method}--\cref{sec:eval:taxonomy}) & Human-verified output from the version in use at the time, before CTC re-scoring, the prior, and the review stops of \cref{sec:recon:dualsheet} & Independent expert record, disagreements adjudicated \\
Crown section, same 45 games, re-recorded & Review stops and absurdity comparison (\cref{sec:eval:ablation}); learned signals (\cref{sec:eval:learned}); engine test (\cref{sec:eval:engine}) & Release v0.12.40; oracle harness with both sheets' cell logits (CTC) and readings; prior applied afterwards. Stored OCR from the original reads (May--June 2026, before the rule cleaning of \cref{sec:arch:grid} existed); two games re-read in August after the fix & Adjudicated record from the expert comparison \\
Second section, 44 of 45 games & Validation of the prior (\cref{sec:eval:learned}); review stops (\cref{sec:eval:ablation}); engine test (\cref{sec:eval:engine}) & As above, no cell logits; prior at its frozen shipped weight. By file date, 25 games were read while the defective pixel erasure of \cref{sec:arch:grid} was in use, 19 after the fix & Operator-verified record \\
Open section of a junior championship, 39 games & Confirmatory engine test (\cref{sec:eval:engine}); out-of-sample test of the learned signals (\cref{sec:eval:learned}); review stops (\cref{sec:eval:ablation}) & Engine test recorded with the code fixed at registration (release v0.12.8) and, as a check, with v0.12.40; learned signals on v0.12.40, with both sheets' cell logits; review stops on both releases & Operator-verified record \\
Adjacent section, 15 games & Measured operator effort (\cref{sec:eval:results}) & Batch mode with effort logging & --- \\
\bottomrule
\end{longtable}

\subsection{Methodology: two independent records and three-way
adjudication}\label{sec:eval:method}

\begin{figure}[t]
\centering
\begin{tikzpicture}[
  rec/.style={draw, rounded corners, align=center, minimum height=11mm,
              text width=28mm, font=\small, fill=blue!4},
  proc/.style={draw, align=center, minimum height=13mm, text width=30mm,
               font=\small, fill=black!4},
  src/.style={draw, rounded corners, align=center, minimum height=13mm,
              text width=30mm, font=\small, fill=green!6},
  result/.style={draw, double, rounded corners, align=center, minimum height=13mm,
              text width=20mm, font=\small, fill=orange!8},
  arr/.style={-{Stealth[length=2.5mm]}, thick}]
  \node[rec] (zug) at (0,0.95) {Record~1\\ \textbf{Zugwise}\\ {\scriptsize human-verified}};
  \node[rec] (exp) at (0,-0.95) {Record~2\\ \textbf{the expert}\\ {\scriptsize independent}};
  \node[proc] (work) at (4.6,0)
        {Programmatic comparison\\ $\rightarrow$ disagreement\\ worklist};
  \node[src] (sheet) at (9.2,0)
        {Original \textbf{scoresheets}\\ {\scriptsize human reviewer, in context}};
  \node[result] (verd) at (13.4,0) {Per-ply\\ \textbf{verdict}};
  \draw[arr] (zug) -- (work);
  \draw[arr] (exp) -- (work);
  \draw[arr] (work) -- (sheet);
  \draw[arr] (sheet) -- (verd);
\end{tikzpicture}
\caption{Three-way adjudication. Neither record is ground truth for the other;
the two independent transcriptions are compared to produce a disagreement
worklist, and every disagreement is adjudicated against the original
scoresheets by a human reviewer reading the handwriting in context. The
comparison is itself a defect-finding instrument---it surfaced, among other
things, an expert transcription that left a queen hanging for several moves.}
\label{fig:adjudication}
\end{figure}
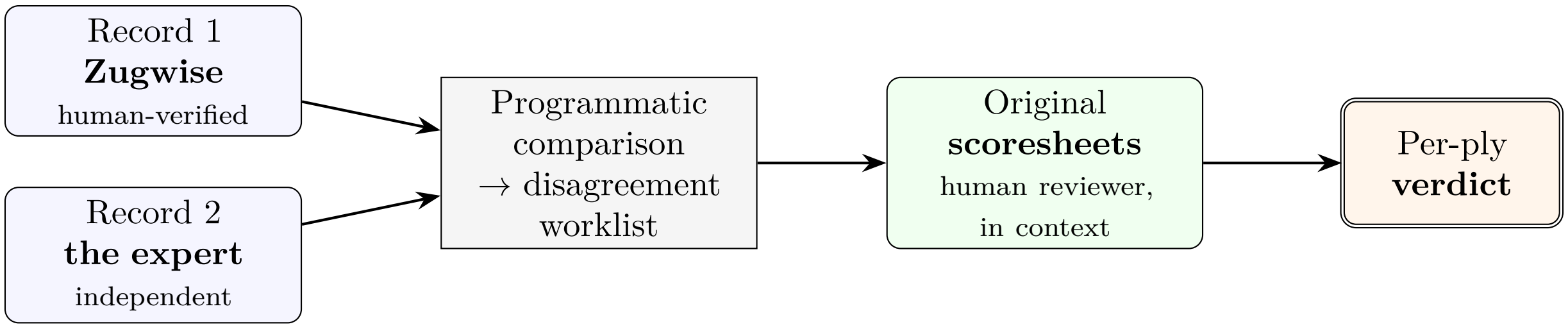

\textbf{Corpus.} One complete round-robin section of a rated
over-the-board tournament: \textbf{45 games} among ten players rated
1678--2167 (mean 1866), two handwritten scoresheets per game (one per
player). The corpus is complete: every
game in the section is evaluated, with no exclusions.

\textbf{Record 1: Zugwise.} The author processed all games through
the full pipeline (\cref{sec:arch}--\cref{sec:recon}): dual-sheet OCR, alignment and merge,
reconstruction, and interactive verification. The output is
\textbf{human-verified, not automatic}. Each game received six to eight
minutes of operator review on average (from two to three minutes on
clean, neat handwriting to substantially longer on degraded or
unconventional sheets), in which the operator resolved the ambiguities
the system surfaced. This caveat applies to the expert comparison below.
What is evaluated is the
system \emph{as designed to be used}: an assistive reconstruction
tool with a human in the loop.

\textbf{Record 2: the expert.} An experienced transcriber
independently produced a manually typed PGN of all games from the same
scoresheets, without access to Zugwise or its output. Both players'
sheets were available to the expert for every game; we did not record
how often the second sheet was consulted. The expert's
self-reported pace is roughly five minutes for a clean 40-move game,
rising to 20--30 minutes on difficult handwriting, with roughly one
sheet in a hundred abandoned as unreadable.

\textbf{Comparison.} The two PGN sets were compared programmatically
(\texttt{compare\_pgns.py}; \cref{fig:adjudication}). Since both inputs are legal, complete
games, the alignment is clean-versus-clean and reliable. The comparison
yields, per game: move-for-move agreement, substituted plies (same ply,
different move), and one-sided plies (present in one record, absent in
the other).

\textbf{The comparison as a defect-finding instrument.} An independent
record turned out to surface more than transcription differences. During
adjudication, a small number of disagreements traced not to
reconstruction at all but to defects in the application's persistence
and export layer. Previously truncated trailing moves reappeared in
an export, and an autosave wrote a stale earlier movelist rather than
the final verified one. These defects did not involve the
reconstruction, scoring, or verification logic of \cref{sec:recon}. They corrupted the
\emph{serialization} of games the operator had already resolved. They
were fixed, and the affected games were re-processed through the
corrected pipeline under a fixed policy: only exports invalidated by
an identified defect were re-run, never games whose readings had merely
been adjudicated wrong. All figures below are computed on the
resulting corpus. We did not record how many games were affected. Nor did we record, for
each, whether its export was regenerated from the saved decisions or the
game was verified again after the disagreement worklist had been seen;
the latter would weaken the independence of that game's record. A later
audit found the stale movelists still in the separately saved per-game
copies of 28 games across both sections. In all 28, the round files held
the final record; these are the files from
which the 45- and 44-game corpora of \cref{sec:eval:ablation} and \cref{sec:eval:learned} were built. The check was made against the expert transcription and the
adjudication for the adjudicated section, and against the scoresheets for
the second. We report these defects for transparency, and
because they are themselves a result. A second, independently produced record
is a powerful end-to-end test of a digitization pipeline. Several of
these defects would have been difficult to detect any other way.

\textbf{Adjudication.} \emph{Neither
record is ground truth for the other.} The expert is not an oracle.
The comparison itself surfaces an expert transcription that hangs a
queen for several moves (game 1.2, \cref{sec:eval:results}). Zugwise's output is not
an oracle either. Every disagreement was therefore adjudicated
three-way, against the original scoresheets, by a human reviewer reading
the handwriting in context. Every disagreement in the corpus received a
verdict; none remain open. Two limitations apply. First,
adjudication covers only the \emph{disagreements}. Moves on which the two
records agree were not independently audited against the sheets, so an
error both records share cannot be excluded. The target is the
best-supported reconstruction from the available records rather than a
certified one. Second, the adjudicator is the author, who is also the
system's operator, so the adjudication reviews, in part, its own
homework. The substitution verdicts rest on legible handwriting read in
game context. The genuinely low-confidence calls are confined to
the degraded endgame tails and are flagged as such where they occur
(\cref{sec:eval:results}). Only an independent second adjudicator can bound the risk
of bias, however, and that is planned for the journal version (\cref{sec:eval:noclaim}).

\subsection{Results}\label{sec:eval:results}

\textbf{Agreement.} Of 45 games, \textbf{34 are move-for-move identical}
between Zugwise's human-verified output and the expert's independent
transcription. Two records produced by entirely different processes,
from the same handwriting, agree completely on three games in four. The
remaining 11 games contain 13 substituted plies and a set of one-sided
plies, every one of which was adjudicated.

\textbf{Substituted plies (13, all adjudicable).} The \textbf{expert is
correct on 7 and Zugwise on 6} (\cref{tab:substitutions}):

\begin{longtable}[]{@{}
  >{\raggedright\arraybackslash}p{(\columnwidth - 8\tabcolsep) * \real{0.20}}
  >{\raggedright\arraybackslash}p{(\columnwidth - 8\tabcolsep) * \real{0.20}}
  >{\raggedright\arraybackslash}p{(\columnwidth - 8\tabcolsep) * \real{0.20}}
  >{\raggedright\arraybackslash}p{(\columnwidth - 8\tabcolsep) * \real{0.20}}
  >{\raggedright\arraybackslash}p{(\columnwidth - 8\tabcolsep) * \real{0.20}}@{}}
\caption{The thirteen substituted plies, each adjudicated against the original
scoresheets. Plies are counted from zero (see \emph{Terminology},
\cref{sec:intro}), so ply 18 is 10.W.}\label{tab:substitutions}\\
\toprule
Game & Ply (move) & Zugwise & Expert & Verdict \\
\midrule
\endfirsthead
\toprule
Game & Ply (move) & Zugwise & Expert & Verdict \\
\midrule
\endhead
1.2 & 18 (10.W) & \textbf{Qe2} & Qd2 & Zugwise --- expert's Qd2 leaves
the queen hanging for several moves \\
1.3 & 64 (33.W) & \textbf{Re8+} & Re7 & Zugwise --- confirmed rook
maneuver \\
1.3 & 66 (34.W) & \textbf{Rb8} & Re8 & Zugwise --- continuation of the
same routing \\
3.2 & 50 (26.W) & Re3 & \textbf{Rgf3} & Expert --- full rook routing
confirms g3→f3 \\
5.1 & 115 (58.B) & \textbf{Kd5} & Kd4 & Zugwise \\
5.1 & 116 (59.W) & \textbf{Kd3} & Kd2 & Zugwise \\
5.1 & 141 (71.B) & Kg6 & \textbf{Ke6} & Expert --- king-walk square \\
5.1 & 153 (77.B) & Kh7 & \textbf{Kf7} & Expert --- king-walk square \\
5.4 & 67 (34.B) & Kg7 & \textbf{Kh7} & Expert --- single-square king
destination \\
5.5 & 49 (25.B) & \textbf{h5} & Ne3 & Zugwise --- two moves missing from
one sheet shifted h5 by a row \\
8.5 & 73 (37.B) & Rc1+ & \textbf{Rc2} & Expert --- voluntary king walk,
not a check escape \\
8.5 & 103 (52.B) & Rg5 & \textbf{Ra3} & Expert --- operator confirmed
before background search resolved \\
8.5 & 145 (73.B) & Kh6 & \textbf{Kh8} & Expert --- operator confirmed on
poor legibility \\
\bottomrule
\end{longtable}

\textbf{One-sided plies (completeness).} Here the asymmetry runs the
other way, and it needs careful framing. The
corpus contains 27 plies present in Zugwise's record but absent from the
expert's. There are 23 trailing plies across four games (1.1, 1.3, 1.4, 2.2) and two
interior plies in game 9.3. Two more plies in game 5.5 come from a dual-sheet
alignment shift at the tail (moves 25--26). There, one scoresheet
recorded a pair of moves the other omitted and the merge carried the
longer reading, a row shift near the end of the game rather than an interior
gap. The trailing plies are almost all endgame technique in drawn or
decided positions, recorded on sheets that visibly degrade under time
pressure. The handwriting collapses toward scribble, and in one game a
player stopped recording altogether. Declining to fight such a tail is a
defensible transcription judgment. At some point any
transcriber reasonably stops on a tail that may no longer be
decipherable at all. Nor did Zugwise read these tails autonomously.
Recovering them required deliberate operator effort, and the
adjudication of these particular plies carries lower confidence than the
substitution verdicts, since the adjudicator is also the operator
(\cref{sec:eval:method}). Put symmetrically, the two workflows make
different choices at the hard tail (one stops, the other digs), and
the archive differs accordingly. The one completeness case that carries
full weight is game 9.3. Two \emph{interior} plies, clearly
present on both scoresheets, are absent from the expert's record and
retained by Zugwise. The omission is two plies, one full
move, and that is what makes it dangerous. Dropping a full move
preserves side-to-move parity, so the expert's shortened record still
parses as a fully legal game wherever the resulting position-jump is
itself legal, as it was here. An interior omission therefore raises no
legality flag, yet it misaligns every move after the gap. Scored against the expert's record, a naive
accuracy metric would miss the two dropped plies and would also count Zugwise's \emph{correct}
continuation as a cascade of errors. The case shows concretely why
neither record can serve as ground truth for the other. In the opposite
direction there is nothing: \textbf{no ply present in the expert's
record is absent from Zugwise's} anywhere in the corpus.

\textbf{Per game.} The ply counts above are not an accuracy comparison,
and the game is the unit of evaluation (\cref{sec:eval:measure}). Zugwise's seven
substitution errors fall in four games (3.2, 5.1, 5.4, 8.5). The
expert's six also fall in four (1.2, 1.3, 5.1, 5.5), with game 5.1 among
both, and the expert's omissions add four more (1.1, 1.4, 2.2, 9.3). Under our
adjudication, including the lower-confidence tails, 41 of 45
Zugwise records and 37 of 45 expert records contain no identified
error. If the tail-only discrepancies of three expert games (1.1, 1.4,
2.2), which we do not count as lapses, are not counted as errors, 40 of 45
expert records have no identified error. These are counts of \emph{identified} errors among disagreements,
not independently established accuracy rates. They do not make
either workflow the more accurate, and they show why a ply-level tally
cannot settle that question either.

\textbf{Effort.} Zugwise's operator cost averaged six to eight minutes
of verification attention per game, with wide variance: clean, neat
handwriting can take two to three minutes, while degraded or unconventional
sheets take considerably longer. These figures are self-reported for the
evaluation corpus. They are consistent with later instrumentation. Per-game effort logging, added after this evaluation,
measured fifteen games of an adjacent section of the same tournament
series. It attributed input-activity time to the open game and discarded
idle gaps longer than two minutes (a conservative lower bound that, for
example, automatically excluded a twenty-minute interruption that fell
mid-review). Measured attention averaged \textbf{5.9 minutes per game}
(median 5.6, range 2.5--13.7), over \textbf{255 logged review decisions}.
These come to seventeen per game, of which 30\% overrode the algorithm's proposal.
The measured distribution is consistent with the lower end of the
self-reported account (a few-minute typical cost with a heavy tail
driven by difficult sheets), though it covers different games, on a
later version of the system. The expert's pace on clean sheets is about
five minutes per game. Majid and Eicher put manual entry at four to six
minutes per game without interruptions and six to eight minutes averaged
over a four-round Swiss tournament, with one to two minutes for their
semi-autonomous mode~\cite{majid2022}; their figures come from event
organizers and from volunteers using their framework. Ours is measured attention during human verification of other games from the same tournament series, on degraded sheets, and it includes the
operator's work on the hard tails. \textbf{We do not establish that either workflow
is faster.} The expert's times and the evaluation corpus's operator times
are self-reported, and the instrumented sample is a different set of
games, so there is no matched comparison. The accounts do suggest a
difference in shape. The expert reports that pace degrades sharply on
difficult handwriting (20--30 minutes, with occasional sheets abandoned
as unreadable, where the system's candidate generation and dual-sheet
evidence still function), while the operator's measured cost has a
heavy but shorter tail. And, as the next subsection shows,
the two processes err on different plies, and only hand transcription drops the hardest moves.

A second component of the operator's work bounds what the candidate
machinery can claim. Most
corrections are cheap: the system stops at a ply or flags a move, and
the correct reading is in the suggestion list. In a minority of
cases, though, the correct move is \emph{neither among the fix suggestions nor at
a ply where reconstruction stopped at all}. A legal-but-wrong reading
played through silently, and the operator detected it only by reading
the game and backtracking manually. The review workload therefore
includes a vigilance component, in which the operator audits the system's
silences as well as adjudicating its questions. The same point follows from the candidate-list bound noted in \cref{sec:recon:fixfinding}:
when the
correct move is in no candidate's list, no ranking can surface it.
For this reason we report operator time as the cost of the whole
\emph{workflow}.

The records compared here were made with a version of the system that
lacked the review stops of \cref{sec:recon:fixfinding} and \cref{sec:recon:dualsheet}. Replayed with
the present system, three of Zugwise's seven adjudicated substitution errors
(\cref{tab:substitutions}) become questions to the operator that the earlier version
never asked. In game 8.5 the present system stops at plies 73 and 103
themselves, and in game 5.4 the hanging-piece check of \cref{sec:recon:absurdity} catches
ply 67 five plies later. The other four were at stops in both versions; at
two of them (game 5.1) neither version offers the correct king square. The
present system thus asks a few more quick questions per game and, in
exchange, stops at almost every error that used to pass silently. Whether
that makes verification faster or slower overall is not measured.

\subsection{Error taxonomy: where the two records err}\label{sec:eval:taxonomy}

Beyond the 7-versus-6 score, the adjudication shows where the errors fall. Among the substitutions, the two records err in the \emph{same} ways, on \emph{different} plies. Every one-sided ply is on one side.

\textbf{Both records misread ambiguous squares and routes.} Six of the seven adjudicated Zugwise substitution errors are OCR character- or destination-square
confusions in \textbf{chess-permissive positions}, where both readings are
legal and both are plausible. In the king walks nothing in the
continuation disambiguates them. King-walk squares in a king-and-pawn
endgame are the canonical case (\texttt{Kg6} versus \texttt{Ke6}, with
either followed by the same recorded continuation). In two others (3.2,
and 8.5 at ply 73) the continuation did carry the answer, as the
verdicts of \cref{tab:substitutions} show, and the verification missed it. These make up the system's
residual error class, the one \cref{sec:recon}
predicts. Where legality, plausibility, and continuation all fail to
discriminate, only the handwriting itself carries the signal. When
the handwriting is ambiguous, the operator can confirm a wrong but
defensible reading. One of the seven (8.5 ply 103) is an operator-pace
slip, a confirmation issued before a background search had resolved.
It is a user-interface timing lesson rather than a reconstruction failure,
but we count it against the system, since the system includes its operator. The expert's substitutions are mostly of the same kind. Two are king-walk squares in game 5.1 (58.B and 59.W), the same game in which Zugwise misread two others (71.B and 77.B). Another is a rook route in game 1.3 that mirrors Zugwise's in 3.2. Each side also has one slip. Zugwise's is the premature confirmation. The expert's is \texttt{Qd2} in game 1.2, which leaves a queen \emph{en prise} for several moves, the kind of error Zugwise's absurdity detection (\cref{sec:recon:absurdity}) exists to catch.

\textbf{Only hand transcription drops moves.} Every one-sided ply is missing from the expert's record. They comprise the hard-tail completeness gap described above (a transcription judgment under degraded, time-pressure handwriting more than an error in any strict sense) and the two interior plies of game 9.3. A further expert-side category is non-substantive but
methodologically interesting: pervasive \textbf{notational
over-specification}. It covers unnecessary disambiguators (\texttt{Nge2} where
\texttt{Ne2} is unambiguous given a pin) and silent normalization of a
player's notation error. These reach the same squares and do not affect
the chess content. They do mean, however, that the expert's record is a \emph{cleaned
interpretation} of the scoresheet.
Human transcription is itself a reconstruction process
with editorial judgment inside it.

Among the adjudicated disagreements, no ply appears in both records'
error lists. That outcome is partly by construction, because a ply both records got
wrong in the same way is an agreement, and agreements were not audited.
It shows that the two processes err on different plies where they differ, not that they never err together. Game 5.1 makes this concrete: each record misreads king-walk squares that the other reads correctly. In game 9.3, the expert's record is missing two correct interior plies that
Zugwise retained. \textbf{At no adjudicated disagreement were both
records wrong}: every time, the sheets supported one of the two readings.
(That a record reconciled from both contains none of the identified
errors follows by construction, since errors were identified only
through disagreements; that each record catches errors the other
contains does not.) A conventional evaluation that anointed
either one as ground truth would have scored the other's \emph{correct}
moves as errors. The expert's reference would score Zugwise's correct
\texttt{Qe2} as a mistake, and vice versa for the king walks.

These findings justify the section's methodology empirically, and
the lesson generalizes. When two transcription processes err on different plies,
a single accuracy number computed against either record is misleading
in both directions. We instead report the stratified disagreement
worklist with three-way adjudication and a taxonomy of errors by source.
The corresponding system-level claim is deliberately modest. Assistive
reconstruction with human verification produced a record of broadly
similar quality to a contemporaneous expert transcription, with errors on different plies where the two disagree. Either process
checked against the other is therefore stronger than either alone.

\subsection{Ground truth and its hazards}\label{sec:eval:groundtruth}

The adjudicated corpus above doubles as the paper's caution about ground
truth. Our internal regression suite is built on a small set of fully
adjudicated games (including the matched pair of \cref{sec:recon:matchedpair}, Games A and B,
whose every ply has a scoresheet-verified verdict). Its development
surfaced an asymmetry: \textbf{a wrong ground truth does
asymmetric damage.} If the reference itself contains an error, then an
OCR pipeline that happens to reproduce that error scores perfectly,
while a pipeline that correctly reads what the player wrote is scored as
broken. The evaluation then actively selects for the wrong behavior. Game
1.2 is the live example. Had we used the expert's record as ground
truth, Zugwise's correct \texttt{Qe2} would have been counted as a
system error, and any tuning against that ``error'' would have damaged
the system. Reference records for this task must themselves be
adjudicated against the primary source.

A second hazard is reporting oracle-assisted measurements as autonomous
performance. Our test harness (\texttt{test\_recorder.py}) replays games
against adjudicated references and records, at each stuck point, whether
the correct fix appears among the system's candidates and at what rank.
The oracle selects the correct candidate (or injects the
reference move when no candidate matches) so that the replay can
continue and \emph{every} stuck point in the game is measured. These
rank metrics characterize the fix machinery's reachability and ranking
quality, and we use them for regression testing and tuning. They are
\textbf{not} an autonomous-recovery rate, and we deliberately do not
report a ``percent solved'' figure from them. The oracle's selections
stand in for the human operator, so such a figure would describe the
oracle's performance.

Read in that light, the rank metrics themselves are informative
about review cost. A \emph{decision point} is a stop of the oracle replay at
which an action is recorded: confirming a flagged move as written, choosing a
replacement, or inserting or deleting a move. Across the eight-game
regression corpus (the same corpus as \cref{sec:eval:ablation}'s ablation and \cref{sec:eval:priorwork}'s
baseline), the replay stops 87 times, 10.9 times per game. At 76 of these
stops the correct action is to repair a misread, and at 8 it is to confirm a
correct reading; of the other three, two are corrections in no candidate
list and one is an inserted move. The count per game depends on the corpus. The 45 games of
the adjudicated section average 41 (1{,}846 in all), because their degraded
tails stop the replay far more often; the 1{,}290 decisions of
\cref{sec:eval:learned} are the repairs with the correct move among the candidates. The
seventeen review decisions per game of \cref{sec:eval:results} are a different quantity:
operator actions logged in batch-mode review, on other games and an earlier
version. At 63 of the 76 repairs (83\%) the correct fix is the system's
top-ranked suggestion, so the operator's action is a confirmation, and at 71
(93\%) it is in the top three. Two further corrections are in no candidate
list at all. These numbers describe reachability and ranking, that is, how cheap
the review is when it works. They do not remove the vigilance
component (\cref{sec:eval:results}). A legal-but-wrong move that never stops the replay is
not a decision point at all, and on the larger corpus of \cref{sec:eval:learned} the
correct move is sometimes in no candidate list.

The harness itself needed the same scrutiny it applies to the system. An
earlier version accepted an oracle insertion whenever the inserted move
appeared in the reference up to two plies later, rather than at its own
ply; in one regression game this inserted a move the player never
omitted, misaligned everything after it, and more than doubled the
game's recorded decision points. All figures in this section and the next were
regenerated after the defect was corrected. The defect is
the ground-truth hazard of the first paragraph in another form. A
measurement instrument that is subtly wrong produces numbers that are
internally consistent and look entirely normal.

\subsection{Ablation: which components carry the load}\label{sec:eval:ablation}

To attribute the system's behavior to its components, we re-ran the
oracle harness of \cref{sec:eval:groundtruth} on the same eight-game corpus with one component
disabled at a time: the absurdity machinery (the quiescence-backed
detection of \cref{sec:recon:absurdity}, including early absurdity detection during replay),
the character-confusion model (the confusion table emptied; the
similarity machinery otherwise intact), the missed-free-capture penalty,
the free-capture-with-check penalty, and Phase 2 backtracking (search
behind the validation frontier, \cref{sec:recon:fixfinding}). Because the eight-game corpus is
small, we repeat the two most important comparisons on all 45 games of the
adjudicated section: with and without the review stops, and with and without
absurdity detection. All figures below are oracle rank metrics under
\cref{sec:eval:groundtruth}'s framing: reachability and ranking, not autonomous recovery.
They count repair decisions, as \cref{sec:eval:learned} does. Disabling a component
changes which decisions occur as well as how they are ranked (76 against 74
repairs, for example), so the percentages in a row are not paired
measurements on an identical set of events; the per-decision comparison
below follows each error individually instead.

\begin{longtable}[]{@{}lllll@{}}
\caption{Ablation on the eight-game oracle corpus: repair decisions with
each component disabled. The last column counts corrections that are in no
candidate list.}\label{tab:ablation}\\
\toprule
Configuration & Repairs & Top-1 & Top-3 & In no list \\
\midrule
\endfirsthead
\toprule
Configuration & Repairs & Top-1 & Top-3 & In no list \\
\midrule
\endhead
Full system & 76 & 63 (83\%) & 71 (93\%) & 2 \\
− absurdity detection & 74 & 58 (78\%) & 68 (92\%) & 3 \\
− character-confusion model & 77 & 62 (81\%) & 72 (94\%) & 3 \\
− missed-free-capture penalty & 76 & 61 (80\%) & 69 (91\%) & 2 \\
− free-capture-with-check penalty & 78 & 66 (85\%) & 73 (94\%) & 2 \\
− Phase 2 backtracking & 70 & 59 (84\%) & 66 (94\%) & 7 \\
\bottomrule
\end{longtable}

\textbf{The review stops catch what used to pass silently.} The review stops
of \cref{sec:recon:fixfinding} and \cref{sec:recon:dualsheet} re-check the confirmed prefix, stop wherever
the two sheets' best legal readings differ, offer the piece another sheet
names, and catch an ignored free capture with check. The present release
cannot switch them off, so we compare it with an earlier release that lacks
them, v0.12.8. That release also scores CTC against one sheet only and lacks
some smaller fixes to the searches. The comparison therefore measures the
combined effect of the changes and does not isolate the review stops. Two
terms need defining. A \emph{silent
error} is a wrong reading that the replay never stops at (a legal-but-wrong
move, or a missing or extra move), which the oracle
repairs without a decision once a later stop shows that the confirmed record
cannot be right. A \emph{wrong move left in the final record}
is one that neither the system nor the oracle repaired. On the adjudicated
section (\cref{tab:stops}), the newer release removes nearly all silent errors
(16 to 1) and cuts the wrong moves left at the end from 10 to 3. On the
second section it cuts silent errors by three quarters (50 to 13). That section has no cell
logits, so the change to CTC scoring cannot explain its result there, though
the other search changes could contribute. Both sections
are within-sample in one respect: the stops were developed while digitizing a
further tournament, but their false stops were checked on these two sections.
That tournament's Open section is the test set of \cref{sec:eval:engine}, and
there the newer release cuts silent errors from 11 to 1 and the wrong moves
left from 9 to 2. These figures cover the 37 games that the harness completes with both
releases: one game reaches the iteration limit, and in the other the replay
ends at a stop with nothing to propose. (The usable games of
\cref{sec:eval:learned} and \cref{sec:eval:engine} leave out a different pair, so
their decision counts are not subsets of these repairs.) Because the stops were developed on
this tournament, its section does not test them independently.
The cost is about a quarter more stops (a third on the Open section), most of them confirmations of a
reading that was already correct. On the adjudicated section that is about
seven extra confirmations per game. Whether that makes verification faster or
slower overall is not measured. The remaining games that differ from the
reference without a wrong move differ only in notation, for example a missing
capture sign (\texttt{Qd4} for \texttt{Qxd4}) or an unnecessary disambiguation
(\texttt{R8c8} for \texttt{Rc8}).

\begin{longtable}[]{@{}llll@{}}
\caption{The review stops on the two sections of \cref{sec:eval:learned} and on the
Open section of \cref{sec:eval:engine}, oracle replay, without (release v0.12.8) and
with the stops. The Open section is the tournament on which the stops were
developed. Other stops are insertions, deletions and corrections in no
candidate list.}\label{tab:stops}\\
\toprule
& Adjudicated (45) & Second (44) & Open (37 of 39) \\
\midrule
\endfirsthead
\toprule
& Adjudicated (45) & Second (44) & Open (37 of 39) \\
\midrule
\endhead
Silent errors & 16 → \textbf{1} & 50 → \textbf{13} & 11 → \textbf{1} \\
Wrong moves left at the end & 10 → 3 & 11 → 6 & 9 → 2 \\
\quad games with such a move & 8 → 3 & 5 → 3 & 8 → 2 \\
Games differing only in notation & 13 → 12 & 9 → 9 & 8 → 2 \\
Stops in all & 1{,}456 → 1{,}846 & 1{,}602 → 1{,}985 & 971 → 1{,}285 \\
\quad confirmations & 56 → 367 & 109 → 438 & 44 → 329 \\
\quad repairs & 1{,}242 → 1{,}322 & 1{,}346 → 1{,}431 & 829 → 874 \\
\quad other & 158 → 157 & 147 → 116 & 98 → 82 \\
\bottomrule
\end{longtable}

\textbf{Absurdity detection stops the replay where the error is.} Removing
it lowers the top-ranked rate on the eight games from 83\% to 78\%, adds a
correction that no list contains, and lets one error play through silently.
The large corpus shows how. On all 45 games the detector stops the replay 57
times. At 22 of these stops (39\%) the flagged move is correct as written and
needs only confirmation, and 2 are resolved by inserting a
move the player omitted. At the remaining 33 the oracle repaired a ply: 30
by a replacement (14 at the flagged ply, 16 at an earlier one found by the
backward search) and 3 by a correction no candidate list contained. Those 33
repaired errors are the population followed below. That they are so few is
the effect of the review stops, which reach most errors the detector would
catch first, at the error itself. The full system stops a median of zero
plies from them and offers the correct fix at 30 (17 ranked first). With the
detector disabled, the same 33 errors fare as follows:

\begin{itemize}
\tightlist
\item
  \textbf{1} still stops at the error itself;
\item
  \textbf{24} (73\%) surface \emph{later}, a median of two plies after the
  error (mean 3.6, at most 12). The operator is then shown a stop at a move
  that is correct as written, and the cause lies elsewhere;
\item
  \textbf{4} (12\%) are never flagged at all and remain wrong in the final
  record despite the oracle;
\item
  4 cannot be classified cleanly (for example, absorbed by a structural
  edit).
\end{itemize}

\noindent (All figures in this comparison include CTC re-scoring but not the
prior, which the harness does not load.) Across the section, disabling the
detector removes 26 of 1{,}846 decision points, costs 161 top-ranked fixes
(943 to 782), raises the oracle's silent repairs from 1 to 5, and lowers mean
accuracy under the oracle only from 99.1\% to 98.9\%. The harness repairs
most of what the engine misses, so accuracy says little about this
component. It is better judged by where it sends the operator. For most of
the errors it catches, the detector's contribution is that the stop comes at
the error rather than plies later, where symptom and cause are separated;
for about one error in eight, nothing else would have stopped the replay at
all. It is also the expensive component: disabling it cuts the eight-game
corpus's runtime by a factor of about 4.6 (473\,s to 102\,s), the cost the
hybrid two-layer design of \cref{sec:recon:absurdity} exists to contain. (These are
wall-clock times of the oracle harness in CPython 3.13 on one laptop, an
Intel Core Ultra 7 268V with 32\,GB, one game at a time, measured inside the
recorder and so excluding process start-up; the browser runs the same Python
under Pyodide, more slowly.)

\textbf{Backtracking recovers the cause from a late symptom.} The same
comparison measures the backward search of \cref{sec:recon:fixfinding} directly. For every
one of the 24 errors that surfaced only plies later, the search went back
past the symptom and offered the correct earlier fix; the oracle never had to
supply it. It ranked that fix first for 12 of them (50\%) and in the top
three for 19, close to the full system's rate among the detector-triggered
repairs (17 of 33, 52\%), from a stop a median of two plies away. Detection
and backtracking are thus complementary layers: the
first decides where the replay stops, the second makes the reconstruction
robust when it stops in the wrong place. (The oracle has already repaired
every error it met at earlier stops, so each late stop is measured with the
record correct up to the error being traced; this measures the search on one
error at a time, and in unassisted use several nearby errors can compound.)

\textbf{Phase 2 is what reaches behind the frontier.} Without it, five more
corrections fall outside every candidate list (7 against 2), all of them
where Phase 2 is needed: in one game, corrections whose root cause lies
\emph{behind} the validation frontier become unreachable (16 of 16 in a
candidate list with Phase 2, 10 of 15 without). The symptom appears plies
after the cause, and the cause has already been confirmed.

\textbf{The remaining components are individually near-neutral on this
corpus, and we report that plainly.} Removing the character-confusion
model changes no game's accuracy and moves the top-ranked fix by one in
three games, two down and one up: at decision points, ranking is
carried primarily by reach and the quiescence-backed signals, with
confusion similarity as a tiebreaker. Removing the missed-free-capture
penalty costs two top-ranked fixes; removing the free-capture-with-check
penalty gains three. We keep both because they
are nearly free (quiescence runs only on flagged captures), they are
principled members of the permitted-signal family (\cref{sec:recon:scoring}), and the
classes they guard --- a candidate ignoring a genuinely free queen, a
candidate allowing a free capture with check --- are costly when they
occur, as the Game A analysis of \cref{sec:recon:matchedpair} illustrates; but on this corpus their aggregate contribution is within
noise, and we do not claim otherwise.

\subsection{The learned signals: effect on ranking}\label{sec:eval:learned}

The two learned signals of \cref{sec:recon:learned} were added after the
eight-game regression corpus had served its purpose, and a ranking
signal needs more decisions than that corpus holds. We therefore
measure them on the adjudicated 45-game section itself. Each game is
replayed through the oracle harness of \cref{sec:eval:groundtruth} against its adjudicated
record, with the recognizer's stored output and both sheets' cell logits as
input; all 45 games are measured. A \emph{replacement decision} is a stop at which the correct action is to
replace one move and the correct move is among the recorded candidates. The
correct move is the candidate the oracle selects at that stop, identified by
its ply as well as its notation. It is \emph{ranked first} when its score $S(h)$
is the highest in the list, with ties broken by the order in which the
candidates were recorded. The \emph{top-rank count} is the number of decisions
at which the correct move is ranked first, and \emph{top-rank accuracy} is that
count divided by the number of decisions; we give the count, with the
accuracy in parentheses.
There are 1{,}290 such decisions. These
figures are not comparable with \cref{tab:ablation}: the corpus is harder and
larger, and every replacement decision is counted, including those in
degraded endgame tails.

The two weights were set at different times. The CTC weight was chosen once,
by leave-one-round-out cross-validation on an early recording of this
section, and has been frozen since. The CTC row is therefore not an
out-of-sample measurement on this section; the Open section below provides
one. Here CTC is fixed at that shipped value, and its contribution is
removed exactly by subtracting its recorded term (the recordings store the
score and both learned terms at full precision). Repeating the choice
on the present recording, leave one round out, would select a somewhat higher
weight, and the fold-selected weights rank 900 fixes first against 902 at the
frozen weight: the frozen value costs nothing measurable. The prior's weight
is chosen in the table by leave-one-round-out cross-validation: for each of
the nine rounds, the weight from $\{0, 1, 2, 3, 4, 5, 6, 8, 10, 12, 16, 20,
24, 32\}$ with the highest top-rank accuracy on the other eight rounds (the
smallest such weight on ties) is applied to the held-out round. At the shipped weight instead, never refitted, the
combined signals reach 1{,}010 (78.3\%), similar to the fold-selected result.
The result is insensitive to the choice of weight, which is not the same as
being free of tuning. Because this section also informed the CTC-weight
selection and system development, these are within-corpus evaluations rather
than a fully independent test of the complete pipeline; the second and the
Open sections below are the independent ones. Decisions from the same game are not
independent, so all tests below are run over \emph{games}. For two
configurations $A$ and $B$, let $a_g$ and $b_g$ be the numbers of decisions in
game $g$ at which the correct move is ranked first. The sign test is the exact
two-sided binomial test, with probability one half, of the number of games
with $b_g > a_g$ among the games with $b_g \neq a_g$; tied games are left out,
and $p = 1$ if every game is tied. For the interval, let $G$ be the number of
games and $n_g$ the number of decisions in game $g$. Each of 10{,}000
replicates (fixed seed) draws game indices $I_1, \ldots, I_G$ independently
and uniformly from the $G$ games, with replacement, and computes
\[
\widehat{\Delta} \;=\; 100 \, \frac{\sum_{i=1}^{G} \bigl(b_{I_i} - a_{I_i}\bigr)}{\sum_{i=1}^{G} n_{I_i}}
\]
in points. A game drawn twice counts twice in both sums, and both
configurations use the same draws, so the comparison stays paired. The
per-decision outcomes are held fixed: cross-validated weights are not
refitted within a replicate. The 95\% interval runs from the 2.5th to the
97.5th percentile of the 10{,}000 values. Recurring players within one
tournament make even these intervals a statement about this corpus, not
about new tournaments.

\begin{longtable}[]{@{}lrrr@{}}
\caption{Top-rank accuracy of the correct fix on the adjudicated
section (1{,}290 decisions, 45 games). ``Low similarity'': the correct
move's character similarity to the OCR read is below 0.6 (394
decisions).}\label{tab:learned}\\
\toprule
Scoring & Top-1 & Top-1, low similarity & Top-3 \\
\midrule
\endfirsthead
\toprule
Scoring & Top-1 & Top-1, low similarity & Top-3 \\
\midrule
\endhead
Hand-built signals only & 758 (58.8\%) & 116 (29\%) & 1{,}046 (81.1\%) \\
\quad + CTC re-scoring & 902 (69.9\%) & 195 (49\%) & 1{,}111 (86.1\%) \\
\quad + human-move prior & 924 (71.6\%) & 201 (51\%) & 1{,}179 (91.4\%) \\
\quad + both & \textbf{1{,}006 (78.0\%)} & \textbf{257 (65\%)} & \textbf{1{,}210 (93.8\%)} \\
Prior alone & 557 (43.2\%) & 168 (43\%) & 973 (75.4\%) \\
\bottomrule
\end{longtable}

Three results stand out (\cref{tab:learned}). First, \textbf{each signal is
worth far more than all re-weighting of the hand-built signals combined}
(\cref{sec:recon:fixfinding}). CTC re-scoring adds 144 top-ranked fixes (11.2 points; 95\%
interval 7.7--14.7), and the prior 166 (12.9 points; 10.1--15.5), against
3.1 points for the best linear re-weighting of the hand-built signals that we
tested on the same decisions. Both gains hold at the game level. The prior improves 35 games and
worsens one ($p < 10^{-8}$); CTC re-scoring improves 28 and worsens 3
($p < 10^{-5}$). Together they gain 19.2 points (15.3--22.8; 36 games
better, 2 worse; $p < 10^{-8}$). Second, \textbf{the two signals add up.}
Together they add 248 top-ranked fixes, 80\% of the sum of their separate
gains, and each still helps when the other is present. Once the prior is in
place, CTC re-scoring adds a further 82 (6.4 points; 3.7--9.1; 29 games
better, 9 worse; $p = 0.002$); once CTC is in place, the prior adds a further
104 (8.1 points; 5.1--11.1; 29 games better, 9 worse; $p = 0.002$). Re-reading the ink and knowing how
people play are different evidence, and the scorer uses both. Third,
\textbf{they act where the hand-built score is weakest}. When the correct move
looks little like the OCR read, character similarity cannot find it. There,
the hand-built score ranks it first 29\% of the time, and the two learned
signals together raise that to 65\%. The prior alone is a worse ranker than
the hand-built score overall (43.2\% against 58.8\%), yet in this bucket it
beats it outright (168 against 116). That is the complementarity the additive
design depends on. With the prior, top-10 visibility on the adjudicated
section rises from 97.9\% to 99.8\%.

That last figure is conditional, and it should not be read as a promise of
automatic reconstruction. It counts only decisions at which the correct move
is already among the engine's candidates. Over every decision point in the
section, including structural repairs and the decisions at which the correct
move appears in no candidate list at all, the correct action is among the
engine's top ten at 1{,}641 of 1{,}846 (88.9\%). Along the recorded
oracle-assisted trajectories, 43 of the 45 games required at least one
correction outside the available candidate lists. A search over the top ten
candidates of those lists would therefore reconstruct almost none of these
games on its own, whatever its running time: this is the per-decision versus
per-game arithmetic of \cref{sec:eval:measure} once more. What closes the gap is the
operator, who supplies the moves no list contains.

\textbf{Validation on a second section.} The weights were chosen on the
adjudicated section, so we also measure a second round-robin section that
played no part in choosing them: ten players rated 1485--1825 (mean 1606), 44
of its 45 games, 1{,}326 decisions. (The forty-fifth, whose sheet had been read
with the wrong column layout, was completed from a transcription made outside
the system, so it is not part of this reference.) Its reference is the
operator-verified record rather than an independently adjudicated
transcription, so it may retain errors or preferences of the workflow that
produced it, which lacked the prior; the direction and size of any such bias
are unknown. This section was also recorded without cell logits, so only the
prior can be measured there. With the prior at its shipped weight, frozen and
not refitted, top-rank accuracy rises from 765 (57.7\%) to 916 (69.1\%): 11.4
points (95\% interval 8.5--14.2), with 32 games better and 4 worse
($p < 10^{-5}$). Re-fitting the weight by cross-validation on this section
gives somewhat more (932, 70.3\%), with every fold choosing a slightly lower
weight than the shipped one.

\textbf{Out-of-sample test on a third section.} Both weights were fixed before
the Open section of \cref{sec:eval:engine} was played, and that section was recorded
with both sheets' cell logits. It therefore tests both signals at their
shipped weights, neither refitted. Recorded with the present release, its 37
usable games give 885 replacement decisions (one game has no replacement
decision, and one reaches the harness's iteration limit). CTC alone raises top-rank
accuracy from 538 (60.8\%) to 655 (74.0\%): 13.2 points (95\% interval
7.5--20.1), with 24 games better and 1 worse ($p < 10^{-5}$). On top of the
prior it still adds 9.7 points (6.3--13.2; 23 better, 1 worse). The prior adds
10.1 points without CTC (5.5--14.7; 20 better, 6 worse) and 6.6 points on top
of it (4.4--8.8; 20 better, none worse). Together the two signals reach 713
(80.6\%), 19.8 points above neither (13.9--26.2). The test is cleaner for the
prior than for CTC. The prior was unchanged throughout, but CTC's present
form, scored against both sheets, was introduced while this section was being
digitized, so only CTC's weight was fixed in advance. The reference is again
the operator-verified record, made with the system during the tournament, so
the caution stated for the second section applies here as well.

\textbf{Cost on correctly recorded blunders.} A ranking signal could gain on
average and still bury the moves players really blundered. Using the engine
evaluations of \cref{sec:eval:engine} (the Lite build at 20{,}000 nodes), we
checked the decisions whose written move lost at least 300 centipawns, in
the sense of $L(h)$ defined there. On the second section the correct move is in
the top three at 22 of 32 such decisions with the prior and at 21 without it, and
first at 10 against 11. On the adjudicated section the two signals together
raise the top-three count from 27 to 31 of 38, and on the Open section
(present recording) from 9 to 18 of 19. In these small subsets the signals did not reduce aggregate
top-three retention, although individual correctly recorded blunders can
still be demoted.

\textbf{Move-prediction accuracy is not the target.} A variant prior trained
only on games with at least ten minutes of base time, closer to
over-the-board play, predicted held-out moves 1.4 points more often. It made
no meaningful difference on either section (one more top-ranked fix on
each). A prior should be selected by
the reconstruction measurement, not by how often it predicts the move
played.

\textbf{Scope.} These are oracle rank metrics under the framing of
\cref{sec:eval:groundtruth}. They measure how often the operator's action at a decision is a
confirmation rather than a search. They say nothing about autonomous
recovery, and the adjudicated comparison of \cref{sec:eval:results} was carried out
\emph{before} either signal existed. The prior and CTC re-scoring may reduce
the operator's candidate-selection effort, since the correct fix is first
more often, but that has not yet been measured in operator time or final
error, and they do not change what the operator confirmed.

\subsection{Testing the design rule: a searched engine evaluation}\label{sec:eval:engine}

The design rule of \cref{sec:recon:scoring} keeps general engine evaluation out of
ranking, because it can penalize moves players really made. To test that
choice rather than assert it, we added the strongest available evaluation,
Stockfish~19~\cite{stockfish}, under the protocol used for the prior.

\textbf{Protocol.} For a candidate $h$ with position $b_h$ before its ply
(\cref{sec:recon:learned}), the engine runs two searches on $b_h$, each with one
thread, a 64\,MB hash table, a fresh game and the same node budget, which
applies to the search as a whole: an unrestricted search, whose score is
$V^{\star}(b_h)$, and a search restricted to that ply's candidates as root
moves, with one principal variation per candidate, whose line for $h$ has
score $V_h$. Scores are in centipawns from the point of view of the side to
move in $b_h$; a forced mate in $n$ moves, as the engine reports it, counts as
$10{,}000 - n$ for the mating side and $-(10{,}000 - n)$ for the side being
mated (the conversion of python-chess, which also covers a position that is
already mate). If both scores exist, the candidate's \emph{centipawn loss} is
$L(h)=\max\{0, V^{\star}(b_h)-V_h\}$; the maximum covers
the case where the restricted search scores a candidate above the
unrestricted one. With a weight $\lambda_E \ge 0$ and a cap $C$ in centipawns,
the engine term is $E(h) = -\lambda_E\min\{L(h), C\}/100$, so
$-\lambda_E C/100 \le E(h) \le 0$, and the ranking score becomes
$S(h) + E(h)$, with $S$ as in \cref{sec:recon:learned} and the prior at its
shipped weight. A candidate gets no term ($E(h) = 0$) if its position cannot
be rebuilt, its move does not parse in that position, or the restricted
search returns no line for it. Where the
weight is cross-validated, it is chosen leave-one-round-out from
$\{0, 0.5, 1, 2, 3, 4, 6, 8, 12, 16, 24, 32, 48\}$, preferring the smaller weight
on ties. \Cref{tab:engine} lists the three configurations. We specified the
first test, including its criterion, and committed it to the repository
before running it on the recordings that existed then (release v0.12.8). The
criterion was the game-cluster bootstrap interval: the term counts as helping
on a section if the gain is positive and its 95\% interval excludes zero.

\begin{longtable}[]{@{}llrrl@{}}
\caption{The engine configurations. Loss cap $C$ in centipawns; nodes per
search.}\label{tab:engine}\\
\toprule
Configuration & Engine & Nodes & $C$ & Weight $\lambda_E$ \\
\midrule
\endfirsthead
\toprule
Configuration & Engine & Nodes & $C$ & Weight $\lambda_E$ \\
\midrule
\endhead
Native, cross-validated & Stockfish 19 & 200{,}000 & 1{,}000 & per fold \\
Lite, cross-validated & Stockfish 19 Lite (WebAssembly) & 20{,}000 & 1{,}000 & per fold \\
Fixed in advance & Stockfish 19 Lite (WebAssembly) & 20{,}000 & 100 & 24 \\
\bottomrule
\end{longtable}

\textbf{Cross-validated results.} The first test found a gain on both
sections that met the criterion (+2.4 and +1.8 points), which contradicts the
rule as stated. On the present recordings, with the native configuration, the
gain is somewhat smaller. On the adjudicated section, top-rank accuracy rises
from 1{,}010 to 1{,}030 (+1.6 points; 16 games better, 9 worse, sign test
$p = 0.23$), but the 95\% interval includes zero ($-0.1$ to $+3.3$), so the
criterion is not met there. On the second section it rises from 916 to 938
(+1.7 points; 0.1--3.4; 14 better, 7 worse, $p = 0.19$), which meets it. The
engine weight is cross-validated on both sections here, so neither is a
frozen-weight test of the engine. The engine cannot evaluate 3.9\% and 4.6\% of
the candidates: their position cannot be rebuilt from the recorded move list,
or their move does not parse in it. Since the term is never positive, receiving no term is the best possible
engine score, an implicit reward of the kind \cref{sec:recon:learned} avoids for CTC.
Restricted to the decisions where every candidate is evaluated, the gain is
the same (+1.6 and +1.7 points), so these candidates do not drive it.

The rule's premise survives. Among the repair decisions where the native
engine evaluates the written move, that move has a centipawn loss of at least
100 at 12\% and of at least 300 at about 4\%. Ranked by the engine alone, the correct fix is
first only half the time, less often than with the hand-built score, though
more often than with the prior alone (43.2\%). We call a written move
\emph{sound} when its centipawn loss is under 50. On the adjudicated section,
the term wins 40 decisions and loses none where the written move is sound,
wins 3 and loses 1 between 50 and 99, wins 1 and loses 10 between 100 and 299,
and wins 1 and loses 14 at 300 or more: a net gain of 20. The losses fall where
the player blundered. That is the failure the rule was written to prevent,
now measured rather than anticipated.

Three further observations are exploratory, chosen after the first test was
known. First, a small search suffices. The Lite configuration, small enough
to run offline in a browser, gains about as much as the native one (+1.6 and
+1.4 points). At a very small budget of 100 nodes the gain disappears ($-0.5$
and $-0.1$), but at that budget the engine leaves many more candidates
unevaluated, so this point says little. Second, a cap on the loss bounds the
term's influence. With $C = 100$ and $\lambda_E = 24$, the term lies between
$-24$ and $0$, so it can change the score difference between two candidates
by at most 24 points: a candidate that leads another by more than 24 points
cannot be overtaken because of the engine. That is a bound, not a guarantee
that a correctly recorded blunder keeps its rank, so we measured it. With the
Lite build (whose search finds fewer such losses than the native one) and a
100-centipawn cap on the second section, with the weight chosen per fold (16
in six of the nine folds), the correct move stays in the top three at 22 of
the 32 decisions whose written move had a centipawn loss of 300 or more, as
without the engine, while the term still adds 22 top-ranked fixes. Third, the hand-built quality signals the rule
removed showed no additional benefit once the prior was present
(\cref{sec:recon:scoring}).

\textbf{A fixed configuration on new games.} We then fixed one configuration
in advance (\cref{tab:engine}, last row) and tested it on new games. The settings,
the outcome measures and the decision rule were committed to the
repository~\cite{zugwiserepo} before the games were played (commits
\texttt{76a32d2}, \texttt{871e421} and \texttt{28c8234}; the first test is
\texttt{f825962}). The cap and weight were chosen on the two sections above, so
those sections cannot confirm it. The weight 24 was the cross-validated
choice for this cap on the earlier (v0.12.8) recording of the second
section, in eight of nine folds; on the adjudicated section the folds chose
32. On the present recordings the fixed configuration gains 1.4 points on
the adjudicated section ($-0.1$ to $+3.0$) and 2.0 on the second ($+0.3$ to
$+4.0$). The test is the Open section of a junior
championship, 17 players rated 1702--2013 (mean 1836), 39 games digitized afterwards with an engine-free version of
the system and recorded with the code fixed at registration. Two games
contribute no decision: one has no replacement decision, and the replay of
the other reaches the harness's iteration limit. That leaves 37 usable games and
835 decisions. Top-rank accuracy rose from 660 to 667 (79.0\% to 79.9\%, +0.8
points; 95\% interval $-0.5$ to $+2.5$; 8 games better, 6 worse, $p = 0.79$).
Under the criterion fixed in advance, the gain is not confirmed, and it does
not count against the term either. These are the figures of the registered
analysis script, which identified the correct fix by its notation; one that
identifies it by the oracle's own choice, ply included (\cref{sec:eval:learned}),
gives 664 to 672 of 830 decisions (+1.0 points; $-0.4$ to $+2.6$; 8 better,
5 worse), again not confirmed. The test had more usable games than the
pre-registered minimum of 20, but that does not make it powerful. Under a
rough normal approximation, an interval of this width implies a standard error of about 0.8 points, so a true gain of
1.4 to 2.0 points, what the fixed configuration shows on the two sections
above, would clear zero only about 45 to 75\% of the time, and a true gain of
one point about a quarter of the time. The reference was produced without the tested engine
term, so that term could not influence it. It was not independently
adjudicated, however, and may retain errors or selection effects of the
workflow that produced it; their effect on the comparison is unknown. Where the written move had a centipawn loss of 300 or more, the
correct move stayed in the top three at 13 of 19 decisions (registered
recording), with and without the term. Recording the same games with the present release instead gives
713 to 722 of 885 decisions (+1.0 points; $-0.3$ to $+2.4$), again not
confirmed. The baseline on this
section was already higher than on the two above, which left less to gain.

The rule therefore needs narrowing rather than abandoning. The hand-built
quality signals we tested showed no additional benefit once a population
prior was present. A searched evaluation may carry a little further
information, but on new games we could not show it, and we have not shipped
it.

\subsection{Comparison with prior work}\label{sec:eval:priorwork}

A direct head-to-head with prior systems is not possible. The corpora
differ, the prior systems are not publicly runnable on our data, and,
more fundamentally, the objectives differ. We therefore confine the
comparison to what the published numbers themselves imply.

Eicher et al.~report 90.1\% MRA autonomous and 97.2\% semi-autonomous
on the HCS dataset~\cite{eicher2021}, improved in the journal version to 94.9\% and
98.4\% (95.6\% and 99.1\% with manual four-corner marking) through more
data, a rare-move set and legality checking~\cite{majid2022}. These are strong
recognition numbers, and the present work would not exist without them.
Our OCR layer builds directly on their architecture and their dataset
(\cref{sec:arch}). In a few sentences of their conclusions, both papers also point
in the direction this paper takes. They suggest post-processing over whole-game paths rather than move by move, a graph-theoretic search for the fewest
corrections yielding an end-to-end valid game (which they expect could
bring accuracy very close to 100\%), and the use of the most-played moves
in a position~\cite{eicher2021,majid2022}. Our reconstruction layer was developed independently
and before we encountered that passage, so the two converged on this
direction independently. The formulation differs in the objective. We find
that a valid game is not enough, because a top-to-bottom \emph{legal} game can
still be the wrong game. Our searches, including the Dijkstra
search of \cref{sec:recon:fixfinding}, therefore minimize a cost
built from plausibility and recognition evidence rather than by the number of corrections alone, and they leave the final
choice to the operator. The human-move prior of
\cref{sec:recon:learned} is likewise a concrete, evaluated form of their third suggestion.
Translated to the game level by the arithmetic of \cref{sec:eval:measure},
however, \textasciitilde99\% MRA still expects nearly one erroneous move
per 80-ply game, and the protocol determines \emph{which} errors
survive. Their semi-autonomous mode classifies each sheet's prediction
as \emph{certain} when it is both confident (above a confidence
threshold) and valid notation. It prompts the user only on cross-sheet
conflicts or mutual uncertainty. When one sheet's reading is certain and
the other's is not, the certain reading is accepted \emph{without}
prompting, a path Majid and Eicher themselves identify as a known error source~\cite{majid2022} (their count concerns the two players writing different moves, scored against the handwriting, so it is not by itself a count of legal-but-wrong readings). The residual errors thus concentrate, by construction,
in the class the recognition layer cannot be asked to see. These are
readings confident enough, well-formed enough, and (with the journal
version's legality check) legal enough to pass every automated filter
while still being wrong. Such errors mark the boundary of what any per-box measure can express, and they are no shortcoming of the recognition
work. A
99\% figure reached with human assistance does not mean the recognition
problem is finished. Under such a protocol, however, the errors that remain
are disproportionately the game-level residue, the part this paper
takes up.

The comparison we can responsibly state is therefore architectural,
anchored by the adjudicated corpus and by the legality-only baseline
below. Zugwise surfaced its candidates for human review. All but one of its seven errors were
ambiguous handwriting confirmed by the operator, and the system's role is to
make that review cheap and targeted. It retained hard-tail and interior
plies that manual transcription had, mostly reasonably, let go (\cref{sec:eval:results}).
And the comparison between the two records caught a hanging queen in
the human reference.
That is the contribution the game-reconstruction framing buys.

The claim can also be made quantitative. We used our eight-game adjudicated
regression corpus: games with recorded OCR input and
scoresheet-verified references, six single-sheet and two dual-sheet,
Games A and B among them. On it we replayed the \emph{same OCR input} under
two legality-only protocols spanning the charitable range of prior work.
The \textbf{autonomous} protocol (play the top OCR read if legal in the
running position, else the first legal alternative, else stop)
reconstructs \textbf{zero of the eight games}, with per-game move
accuracy ranging from 2\% to 95\%. The spread \emph{is} cascading
corruption in action. After the first legal-but-wrong read, every later
read is judged against a wrong position, and how far the damage spreads depends largely on where that first read lands (a low accuracy can also reflect the protocol stopping early). The
\textbf{semi-autonomous} protocol is an idealized legality-triggered
baseline. It grants a human who corrects every \emph{illegal} read
perfectly, and it is not a faithful reproduction of earlier systems, which
also prompt on cross-sheet conflicts. It reconstructs \textbf{three of eight}. The other five
games each contain at least one legal-but-wrong read that the protocol
accepts with no signal, the first arriving between ply 7 and ply 48.
From that ply onward the archived game is wrong and nothing in the
pipeline knows it. These five games are \cref{sec:eval:measure}'s silent class measured
directly on real OCR input. They are the games the
reconstruction layer and the verification workflow exist for. With operator
review of the kind whose cost \cref{sec:eval:results} reports on other games, the same
input that defeats both legality-only protocols becomes the
adjudicated record.

\subsection{What this evaluation does not claim}\label{sec:eval:noclaim}

In summary, the scope of the evaluation:

\begin{enumerate}
\def\labelenumi{\arabic{enumi}.}
\tightlist
\item
  \textbf{Nothing here is automatic.} The outputs in the expert
  comparison were human-verified, at six to eight minutes of operator
  attention per game on average (two to three minutes on the cleanest
  sheets). The ranking, ablation and baseline experiments were
  oracle-assisted offline evaluations; they are not further measurements
  of review time or of autonomous recovery. The system reduces and targets human
  effort; it does not eliminate it. Part of that effort is
  auditing the system's silences (\cref{sec:eval:results}) as well as answering its
  questions.
\item
  \textbf{No superiority in speed or accuracy is established.} The
  timings are not a matched comparison, and the adjudicated tallies (7--6
  substitutions to the expert; 41 against 37 games with no identified
  error, or 41 against 40 without the unrecorded tails) count identified errors; they are not accuracy rates.
  The claims are errors on different plies where the records disagree,
  completeness at the hard tails (with the \cref{sec:eval:results} caveats on operator
  effort and adjudication confidence), and the evaluation protocol itself.
\item
  \textbf{Neither record is ground truth, and agreement is not audited.}
  That neither record can serve as the reference is a finding, and it is
  \emph{why} the three-way protocol is needed. That agreeing moves were
  not independently checked is a limitation, so errors both records share
  cannot be excluded.
\item
  \textbf{Scope.} One round-robin section, 45 games, one expert, one
  operator (the author), who also served as adjudicator (\cref{sec:eval:method}). The
  large-scale evaluation (\textasciitilde300 games, 600-plus scoresheets,
  from an open event)
  is deferred to the journal version, as are a second-operator study and
  an independent second adjudicator.
\end{enumerate}

\section{Discussion}\label{sec:disc}

\subsection{When chess knowledge helps and when it hurts}\label{sec:disc:helps}

The matched pair of \cref{sec:recon:matchedpair} shows the limit of position-only plausibility
evidence. In Game A, the free-capture-with-check penalty separates two legal
readings that character similarity cannot, and it points to the reading the
scoresheet confirms. In Game B, the forced-mate penalty penalizes a capture
that walks into mate, yet the player made that capture deliberately, as a way
of resigning. Both readings hand the opponent an immediate forcing reply that
a careful player would not allow. What separates the two games is whether the
game ends there, and only the record shows that.

The engine experiment of \cref{sec:eval:engine} shows the same structure at corpus scale.
Ranked by the engine alone, the correct fix is first only half the time.
Most written moves are sound, so most of those misses are among sound
candidates, where the engine's preference says little about what was
written. Yet a weighted engine term improves aggregate ranking in
cross-validation, and the decisions it loses are concentrated where the
player really blundered, the moves a tactical judgment would penalize. The relevant question is therefore not whether a
signal encodes chess quality, but whether it adds useful evidence at an
acceptable cost to moves that were really played badly. For the hand-built
quality signals the answer was no once the prior was present. A bounded,
searched evaluation helped in cross-validation but did not meet the admission
standard of \cref{sec:recon:scoring}: at its weight fixed in advance, its gain was
not confirmed on new games. The unchanged prior met it; CTC gained as well, but its present form was
introduced while the test section was being digitized.

The system already reflects this in two ways. First, because even tactically verified plausibility evidence can point the
wrong way, each permitted signal is a finite penalty rather than a
categorical exclusion at the scoring stage, although candidate truncation and
the bounded searches can still discard a penalized alternative
(\cref{sec:recon:fixfinding}). (The exclusion of hand-built quality signals rests on
separate evidence: the \texttt{Rxe3} and \texttt{b3} cases, and the measurement
of \cref{sec:recon:scoring}, where restoring them showed no additional benefit once the
prior was present.) Second, the operator holds context the position-only
signals do not: the continuation, the end of the record, the handwriting
itself. \Cref{sec:eval:taxonomy} shows the same structure empirically: where the
system and the human disagree they err on different plies, and at no adjudicated disagreement were both records wrong.

The tension can be narrowed. The fact that separates Game B from Game A is
present in the record: the scoresheet ends at the mate. The forced-mate
signal already suppresses its penalty when every legal \emph{alternative} to
the candidate also allows the mate, that is, when the mate is unavoidable.
The full inversion, which treats an opponent's immediate mate as
\emph{confirmation} when the scoresheet records that mating reply as its last
move, is implementable and is a targeted extension motivated by this example
(\cref{sec:disc:future}). It would narrow the boundary without abolishing it. Games
abandoned mid-position, resignations without a forcing sequence, and time
forfeits will still end records at plies no positional reasoning can
certify, and the operator will remain the right arbiter there.

\subsection{Limitations}\label{sec:disc:limits}

\textbf{Notation scope.} The recognizer and grammar target English
standard algebraic notation. Continental piece-letter conventions are
unsupported, and the failure mode is systematic confusion rather than
graceful degradation. Italian notation, encountered among the sheets we have
processed, uses R for the king (\emph{Re}), which collides head-on with the
English rook. It also uses C, A, T, and D for knight, bishop, rook, and
queen. A sheet written in another notation language is currently at or
beyond the hard end of the operator-effort range. Supporting it properly
requires both grammar localization and retraining, not a
character-mapping patch (\cref{sec:disc:future}).

\textbf{The candidate-list bound.} Reconstruction operates over the
hypotheses the system generates: the recognizer's candidates, the
constrained re-reading over legal moves, and the repair and insertion
hypotheses of \cref{sec:recon:fixfinding}. When the correct move is among none of them, no
search algorithm (greedy, beam, or Dijkstra) can surface it. The
consequence depends on the case. Where the recorded move is illegal, the
stop is flagged and the operator supplies the move. Where a
legal-but-wrong reading plays through, the failure is silent. The
operator's review therefore includes a vigilance component
(\cref{sec:eval:results}): auditing the system's silences as well as
answering its questions. Raising raw recognition quality shrinks
this class but cannot eliminate it. Every system built on a recognition
front-end shares this bound, ours included.

\textbf{Human-in-the-loop by design, and in the evaluation.} Nothing
in \cref{sec:eval} is automatic. The expert comparison reflects six to eight minutes
of operator attention per game on average, and the other experiments
are oracle-assisted. We regard this as the correct
design point, but it bounds the claims.
Further, the evaluation has a single operator, who is also the author
and the adjudicator (\cref{sec:eval:method}), and adjudication covered the disagreements
only. The low-confidence calls are confined to the degraded endgame
tails and flagged as such. Only an independent second adjudicator
can bound the risk of bias, however. Both that and a second-operator study
are deferred to the journal version.

\textbf{Corpus scale and statistical scope.} The evaluation is a
descriptive characterization of one round-robin
section (45 games, the complete section, with nothing excluded or
disqualified), supplemented by a matched pair. The ranking measurements add
a second section with an operator-verified reference and the later Open
section of a junior championship. Raw counts accompany
every proportion, and no inferential claim is made about the adjudicated
comparison. The significance tests in the paper concern ranking only: the
game-level sign tests and bootstrap intervals on changes in the number of
top-ranked fixes in \cref{sec:eval:learned} and \cref{sec:eval:engine}. Large-scale
validation on an open tournament (\textasciitilde300 games, 600-plus
scoresheets) is deferred
to the journal version. The descriptive findings (the silent/loud
split, the errors on different plies, the hard-tail behavior) are
structural in character, and we expect them to persist at scale. That
expectation is what the larger corpus exists to test.

\textbf{Domain shift in the recognition layer.} The BiLSTM reaches
82.9\% raw move-recognition accuracy on held-out HCS writers (\cref{sec:arch:ocr}). On
the real club scoresheets of the regression corpus (diverse layouts,
phone photography, a wide range of handwriting), the corpus
\emph{average} barely moves: 84.0\% raw top-read accuracy on the
single-sheet games. The per-game spread is wide, however, with individual
sheets ranging from 94.6\% down to 68.9\%. In our data, domain shift appears
as a heavy tail rather than a degraded mean. The tail games are where a
legality-only protocol fails hardest (\cref{sec:eval:priorwork}) and where
reconstruction and the operator earn their keep. The reconstruction
layer makes that tail survivable; it does not close it. More and more
diverse labeled training data remains the binding constraint on the
recognition front-end.

\textbf{Correlated artifacts and cross-sheet agreement.} Dual-sheet
evidence rests on an independence assumption: two players' records
rarely fail in the same place, so agreement between them is strong
evidence of correctness. For \emph{handwriting} errors the assumption is
plausible, though this evaluation does not test it. But both sheets pass
through the same processing pipeline, and a
systematic pipeline artifact can corrupt both reads identically. Agreement,
the system's strongest trust signal, then endorses the
error. The fringe-remnant defect of \cref{sec:arch:grid} did this. Both players'
scoresheets carry a printed rule at the same cell position, so both
produced the same confident phantom reading (\emph{Nc4} for a written
\emph{c4}). The readings agreed, the move happened to be legal, and the
merge promoted it to its highest-confidence tier with no flag. The error
was silent at every layer built to catch it. Worse, plies on which both sheets agree are locked for the searches by default, so unless absurdity detection flags the move, no search can repair it; only the operator can. The defect was caught by the operator
during routine review and diagnosed by replaying the recognizer offline
on the stored cell images. The repaired pipeline removed the error class
from subsequent rounds. The structural lesson survives the fix:
agreement between records is only as independent as the process that
produced them, and confidence tiers built on cross-record agreement
should discount agreement that a shared artifact could explain.

\textbf{The population behind the prior.} The human-move prior is
trained on online games in one rating band. Our scoresheets come from
over-the-board classical play across a wider range. On the two round-robin
sections, restricting training to slower games made no measurable difference
(\cref{sec:eval:learned}). The prior's gains are also similar on the three sections, whose mean ratings
span some 260 points; the training band itself was not varied. But the
sections are few, and the right population for a prior is an empirical question we
have only begun to answer. The prior can also cost the top rank when a
player made an unusual move (it did so at 48 of 1{,}290 decisions,
\cref{sec:recon:learned}). A player who plays unusual moves more often than most may therefore be
ranked worse than the average player.

\textbf{Structural errors on a single sheet.} Dual-sheet alignment
recovers missing and duplicated moves because the two records rarely
fail in the same place (\cref{sec:recon:dualsheet}, \cref{sec:eval}). When only one sheet exists, a
structural error has no second record to consult. The insertion repair
of \cref{sec:recon:fixfinding} can hypothesize a forgotten ply where the omission produces a
local stuck point, but in general recovery depends on the operator
recognizing the inconsistency. Single-sheet structural repair is
markedly harder than single-sheet substitution repair, and our results
should be read with that asymmetry in mind.

\subsection{Future work}\label{sec:disc:future}

\textbf{The end-of-game inversion.} Implement the resignation-capture
flip motivated in \cref{sec:disc:helps}. When a candidate move allows an immediate
mate and the scoresheet records that mating reply as its last move,
treat the mate as confirmation rather than penalty. The extension is
targeted, motivated by the paper's canonical hard case, and uses
information already present in the record.

\textbf{Per-scribe reliability weighting.} A scoresheet is one writer's
hand throughout, so aggregate OCR confidence over a whole sheet estimates
that writer's legibility. Weighting the two sheets' evidence by
per-scribe reliability should improve merged candidate pools on the unbalanced pairs
where one record is clean and the other degraded. The hard tails of
\cref{sec:eval} suggest such pairs are common. The extreme case is a player
who stops recording. In one regression game, Black stops recording ten
moves before the end, as the Laws permit a player with less than five
minutes left and an increment of less than 30 seconds
(Art.~8.4)~\cite{fide}. White's sheet continues to the mate. The
recognizer reads Black's
remaining boxes as a near-constant alternation of two pawn moves at
middling confidence, and at two plies the merge still prefers that
invented reading over White's genuine one. Reliability here
should be local rather than per sheet. A run of boxes whose readings
repeat with no regard to the position is evidence that the scribe has
stopped, and those boxes should carry no weight in the merge at all.

\textbf{Notation localization, down to the writer.} Per-language move
grammars and piece-letter maps, plus recognizer retraining (or augmentation)
for continental letterforms, with the Italian king/rook collision as the
acceptance test. The same mechanism is needed below the level of a
language, since a single writer can use idiosyncratic piece letters throughout a
sheet. In one game of the second section, one player wrote the queen as a
lowercase ``a'' and the rook as ``r''. The recognizer read many piece
moves as pawn moves, and the correct move was often absent from that sheet's
candidates. The recognizer maps such glyphs onto its own characters
consistently for a given writer (the Italian-notation writers' knight is
read as Q and their bishop as N), so a per-sheet letter map can be learned
from the other sheet's confident readings. On our two sections it restores
many moves to that sheet's own candidates. In a two-sheet game, however, the other
sheet has usually supplied them already, and applying the maps when merging
left the merged candidates' accuracy unchanged. Its value lies where there is
no second sheet, and in re-reading the ink: scoring each candidate against the
cell as that writer would have written it. A letterform the notation-constrained
decoder never emits as a piece letter, such as the lowercase ``a'', needs the
recognizer itself.

\textbf{Alignment under structural inconsistency.} Extend the
Needleman--Wunsch formulation to the cases the current three-tier merge
handles least gracefully: simultaneous structural errors on both sheets
in overlapping regions. There the alignment itself becomes ambiguous and
should keep the alternatives open. The same game shows the
limit of aligning first and reconstructing afterwards. One sheet omitted a
full move, which the alignment can absorb as a gap. The other sheet, though, was
read so poorly that the two candidate pools rarely overlapped, leaving the
alignment little to work with and the merged pool without the correct move.
The game was recovered outside the system, by aligning and replaying together. Competing
alignments were kept open, and the legality of the resulting continuation
chose between them. Folding the alignment into the reconstruction search,
rather than running it as a preprocessing step, is the natural extension.
A first implementation shows its limit, however. On this game it recovered less
than half of the moves, because the correct move was often in neither
sheet's candidates. The search pays only once recognition, or a per-writer
letter map, has put the right candidates there.

\textbf{A prior matched to the population, and eventually to the
player.} Train the human-move prior on over-the-board games at classical
time controls. Select among candidate priors by the reconstruction
measurement of \cref{sec:eval:learned} rather than by move-prediction accuracy.
Large databases of over-the-board games exist; the open question is
whether they reach the club level where most scoresheets are written.
Matching the population more closely is then a matter of conditioning
rather than of training one model per section. A skill-conditioned model
in the manner of Maia-2~\cite{tang2024maia2} takes the rating as an input,
and tournament metadata already supplies each player's rating, which the
batch pipeline reads from the pairing file. The further step is a prior
conditioned on the \emph{player}. Behavioral
stylometry~\cite{mcilroyyoung2021stylometry} is direct evidence that individual
style is recoverable from moves alone. A scoresheet is one writer's record of both players' moves: writer-specific recognition should be conditioned on the scribe, and a player-specific prior on the player to move at each ply. Each player's own decisions recur in their games, and club players recur across rounds, seasons, and
the archives a club accumulates. The same games this tool exists to
digitize are thus the material such a prior would be built from. The idea
is plausible, but it carries a risk. A prior fitted to
a player is one step further from a measurement over a population, and
the conditions of \cref{sec:recon:learned} would have to be met again, weight and all.

\textbf{Beyond re-weighting.} The weights of the hand-built signals
were hand-tuned, and cross-validated re-weighting shows they leave only
a few points on the table (\cref{sec:recon:fixfinding}). The weights of the two learned
signals were set by cross-validation (the CTC weight on an earlier
recording and then frozen, the prior's per fold), and a frozen prior
weight performs as well; both frozen weights also held on a section digitized
after they were fixed (\cref{sec:eval:learned}). Further gains in ranking
are therefore more likely to come from new \emph{evidence} than from
fitting the existing coefficients more carefully. With candidates scored
against the cells of both sheets, CTC re-scoring now adds
significantly on top of the prior (\cref{sec:eval:learned}). Any learned combination remains
constrained to permitted signals, leaving the scoring design rule
intact.

\textbf{A bounded engine term, revisited.} The engine term of
\cref{sec:eval:engine} helped in cross-validation but was not confirmed on new
games, where its estimated gain was under one point. Whether a different
bound, a stronger search or a section with a lower baseline would change
that is open. Any
such variant would need its own test, fixed in advance, on games not used to
choose it.

\textbf{Scale, independence, and instrumentation.} The
\textasciitilde300-game open-tournament evaluation, and an independent
second adjudicator and a second-operator study. Extension of the
per-game effort instrumentation introduced late in this work (\cref{sec:eval:results}'s
measured attention and per-decision logs, so far covering fifteen games)
to a full corpus, where the per-ply action log becomes reportable
interaction-density and OCR-versus-final confusion data. Growth of
the labeled training corpus, for which the HCS dataset's maintainers are
the natural partners.

\section{Conclusion}\label{sec:concl}

We have argued that digitizing handwritten chess scoresheets is not a
recognition problem with a legality check appended, but a
\emph{reconstruction} problem. The object to be recovered is the game. The
errors that matter most are the ones recognition cannot see: confident
legal-but-wrong readings that pass every character-level, confidence, and
validity filter and can silently corrupt the positions that follow. High
per-move accuracy does not certify a correct game. In pipelines that correct
illegal readings but accept legal ones, the silent readings are the ones that
escape review, and that is where this paper begins.

Our answer is a reconstruction layer that borrows the machinery of computer
chess (quiescence search, exchange evaluation, pin detection) and repurposes
it to ask not ``what should be played?'' but ``could a human plausibly have
written this?'' The architecture enforces the distinction. A design rule
excludes chess-quality signals from candidate ranking, and among hand-built
chess signals it admits only plausibility evidence checked by tactical
search. Added learned and engine-derived terms are held to an admission
standard: at settings fixed in advance, they must improve ranking on games
not used to choose them, with their effect on correctly recorded blunders
reported. Hand-written judgments of the
position showed no additional benefit once the prior was present. A bounded
engine term helped in cross-validation, but its gain was not confirmed on new
games. Detection and repair are separate layers. The absurdity
detector stops the replay at the error, and when an error surfaces only plies
later, frontier-based backtracking traces the symptom back to its cause.
Review stops send every disagreement between the sheets' legal readings to
the operator and re-check what has been confirmed. In oracle replay of the
adjudicated section, a release that adds them and other changes leaves one
silent error where an earlier release leaves 16, at the price of about seven
additional confirmations per game.

The rule still permits learning how people play. A human-move prior,
trained in the manner of Maia, is admitted because it is learned from a
population of games rather than hand-written as a judgment of the position.
Together with re-reading each candidate against the recognizer's raw output,
it raised the correct fix's top-rank rate from 58.8\% to 78.0\% of the
replacement decisions at which the correct move is a candidate, and both
kept their gain without refitting on a later section, the prior fully out
of sample. Each of the
two signals adds on top of the other, and their combined gain is several
times what the best re-weighting of the hand-built signals we tested
reaches. The approach takes up the direction that
Eicher et al.'s conclusion sketched, post-processing over whole-game paths in
search of a valid game, but with a different objective. A fully legal game
can still be the wrong game, so validity must be supplemented by
plausibility. Plausibility itself has a boundary, and only context outside
the position can settle it.

That boundary is why the system is built around a human, and the evaluation
reflects the design. On a complete tournament
section, with every disagreement adjudicated against the sheets, we do not
establish that either the system or an experienced human transcriber is
faster or more accurate. Where the two records disagree, however, they err on
different plies. At no adjudicated disagreement were both wrong, so each
record catches errors the other contains. Alongside the system, we offer the
three-way adjudication protocol for a domain where no single record is ground
truth.

Zugwise, the implementation that validates the approach, runs entirely in the
browser. It is free, private, and offline-capable, with no operational cost,
and it has been used on real tournaments, including events whose games would
otherwise not have been digitized at all. That is the practical point of the
reframing. The games on club scoresheets are waiting for a workflow that
makes a verified archive cheap enough to exist. We hope the reconstruction framing, the plausibility
boundary, and the evaluation methodology presented here are useful to whoever
carries that work further, ourselves or others.

\section*{Code, models, and data}
The application source is public~\cite{zugwiserepo} under the GNU Affero
General Public License, version~3 or later (release \texttt{v0.12.40}, commit
\texttt{8b79d7e}, the version described and evaluated here; the comparison
without the review stops in \cref{sec:eval:ablation} uses release \texttt{v0.12.8}, commit
\texttt{943ca70}), and the recognizer~\cite{zugwisemodel} and human-move prior~\cite{zugwiseprior} are
published models. The evaluation scripts that regenerate every table from
the recordings, and the anonymized oracle recordings behind them
(recognizer candidates, cell logits, per-decision scores and adjudicated
game records), are available from the author on request. The scoresheet
images are not released, to protect the players' privacy.

\section*{Acknowledgments}
The author thanks Hugh Brodie for independently producing the manual reference
transcription used in the evaluation of \cref{sec:eval}; this evaluation would
not have been possible without his work.

\paragraph{Use of AI assistance.}
AI assistance was used under the author's direction: Claude (Anthropic)
for drafting and copy-editing, and for software development, debugging
and running the evaluation scripts; ChatGPT (OpenAI) for critical review
of the manuscript. The author designed the study, made all
methodological decisions, and takes responsibility for all content.

\bibliographystyle{alpha}
\bibliography{zugwise}

\end{document}